\documentclass[journal]{IEEEtran}

\usepackage{graphics,epsfig,graphicx,color}
\usepackage[cmex10]{amsmath}
\usepackage{amssymb}
\usepackage{mathrsfs,amsfonts}
\usepackage{amsthm,amsxtra}
\usepackage{url}
\usepackage{hyperref}
\usepackage[small,nohug,heads=vee]{diagrams}
\diagramstyle[labelstyle=\scriptstyle]

\graphicspath{{./FIGS/}}

\newtheorem{alg}{Algorithm}

\newcommand{\bb}{{\bf b}}
\newcommand{\bc}{{\bf c}}
\newcommand{\bff}{{\bf f}}
\newcommand{\br}{{\bf r}}
\newcommand{\bs}{{\bf s}}
\newcommand{\bt}{{\bf t}}
\newcommand{\bT}{{\bf T}}
\newcommand{\bu}{{\bf u}}
\newcommand{\bw}{{\bf w}}

\title{Automated polyp detection in colon capsule endoscopy}

\author{Alexander V. Mamonov, 
Isabel N. Figueiredo,
Pedro N. Figueiredo,
Yen-Hsi Richard Tsai%
\thanks{A. V. Mamonov is with the
Institute for Computational Engineering and Sciences (ICES),
The University of Texas at Austin, 201 East 24th St. Stop C0200, Austin, 
TX 78712-1229 USA
(mamonov@ices.utexas.edu)}%
\thanks{I. N. Figueiredo is with the
CMUC, Department of Mathematics, University of Coimbra, 
3001-454 Coimbra, Portugal (isabelf@mat.uc.pt)}%
\thanks{P. N. Figueiredo is with the
Department of Gastroenterology, University Hospital of Coimbra 
and Faculty of Medicine, University of Coimbra, 3000-075 Coimbra, Portugal 
(pnf11@sapo.pt)}%
\thanks{Y.-H. R. Tsai is with the
Department of Mathematics and ICES, The University of Texas
at Austin, 1 University Station C1200, Austin, TX 78712 USA
(ytsai@math.utexas.edu)}}

\begin{document}

\maketitle

\begin{abstract}
Colorectal polyps are important precursors to colon cancer, a major health 
problem. Colon capsule endoscopy (CCE) is a safe and minimally invasive 
examination procedure, in which the images of the intestine are obtained via 
digital cameras on board of a small capsule ingested by a patient. The video
sequence is then analyzed for the presence of polyps. We propose an algorithm 
that relieves the labor of a human operator analyzing the frames in the video 
sequence. The algorithm acts as a binary classifier, which labels the frame
as either containing polyps or not, based on the geometrical analysis and 
the texture content of the frame. The geometrical analysis is based on a 
segmentation of an image with the help of a mid-pass filter. The features
extracted by the segmentation procedure are classified according to an 
assumption that the polyps are characterized as protrusions that are mostly
round in shape. Thus, we use a best fit ball radius as a decision parameter
of a binary classifier. We present a statistical study of the performance
of our approach on a data set containing over $18,900$ frames from the 
endoscopic video sequences of five adult patients. The algorithm demonstrates 
a solid performance, achieving $47\%$ sensitivity per frame and over $81\%$ 
sensitivity per polyp at a specificity level of $90\%$. On average, with a 
video sequence length of $3747$ frames, only $367$ false positive frames need 
to be inspected by a human operator.
\end{abstract}



\section{Introduction}

Colorectal cancer is the second most common cause of cancer in women and the
third most common cause in men \cite{jemal2011global}, with the mortality reaching 
to about $50\%$ of the incidence. Colorectal polyps are important precursors to
colon cancer, which may develop if the polyps are left untreated. Colon capsule 
endoscopy (CCE) \cite{adler2003wireless, delvaux2008capsule, eliakim2010video, 
gerber2007capsule, iddan2000wireless, moglia2008recent, moglia2009capsule, 
nakamura2008capsule, spada2010meta} is a feasible alternative to conventional 
examination methods, such as the colonoscopy or computed tomography (CT) colonography 
\cite{eliakim2009prospective}. 

In CCE a small imaging device, a capsule, is ingested by the patient. As the 
capsule passes through the patient's gastrointestinal tract, it records the 
digital images of the surroundings by means of an on-board camera (or multiple 
cameras). As the images are recorded, they are transmitted wirelessly to a 
recording device carried by the patient. Depending on the model of the capsule
and its regime of operation, the images are captured at a rate ranging from 
2 to 30 or more frames per second, with the low frame rate devices being 
prevalent currently. After the whole video sequence is recorded, 
it has to be analyzed for the presence of polyps. The video sequence of 
examination of a single patient may contain thousands of frames, which makes 
manual analysis of all frames a burdensome task. Using an automated procedure
for detecting the presence of polyps in the frames can greatly reduce such 
burden. Thus, an efficient algorithm should not only be able to detect the 
polyps accurately (high sensitivity), but should also have a reasonably low
rate of false positive detections (high specificity) to minimize the number
of frames that have to be analyzed manually.

In this paper we provide an efficient algorithm for detecting polyps in 
CCE video frame sequences. The performance of the algorithm is assessed on a
relatively diverse data set, which ensures that no over-fitting takes place. 
The paper is organized as follows. The main idea and its comparison to existing approaches 
is discussed in section \ref{sec:binclasspre}. The steps of the algorithm 
are described in detail in section \ref{sec:alg}, which concludes with a 
summary of the algorithm. We test our algorithm on a data set comprised of 
the frames from the endoscopic video sequences of five adult patients.
In section \ref{sec:method} the data set and testing methodology are presented.
It is followed by the results of the testing in section \ref{sec:results}.
Finally, we conclude with the discussion of the results and some 
directions for future research in section \ref{sec:discuss}.

\section{Binary classifier with pre-selection}
\label{sec:binclasspre}

The proposed algorithm of polyp detection is based on extracting certain geometric
information from the frames captured by the capsule endoscope's camera. Such 
approach is not new, as it has been noticed before that the polyps can be 
characterized as protrusions from the surrounding mucosal tissue 
\cite{figueiredo2011automatic, liedlgruber2011computer, summers2001automated, 
van2010detection, yao2004colonic}, which was used in CT colonography
and in the analysis of conventional colonoscopy videos \cite{cao2004parsing,
cao2007computer, liu2013optical, liu2013robust, oh2007informative}.
Thus, it is natural to compute some 
measure of protrusion and try to detect the frames containing polyps as the ones 
for which such measure is high. However, this leads to an issue that was also 
observed in the above mentioned works. The issue is distinguishing between the 
protrusions that are polyps and the numerous folds of healthy mucosal tissue. 
This problem can be alleviated by some form of image segmentation that takes 
place prior to the computation of the measure of protrusion 
\cite{condessa2012segmentation}, which is what we do in this work as well.

A particular choice of a measure of protrusion is of crucial importance. Many
authors have proposed the use of principal curvatures and the related quantities,
such as the shape index and curvedness \cite{yoshida2001three}, or the Gaussian
and mean curvatures \cite{figueiredo2011automatic}. The main disadvantage of 
such approaches is that the computation of the curvatures is based on 
differentiation of the image, which must be approximated by finite differences.
In the presence of noise these computations are rather unstable, which requires
some form of smoothing to be applied to the image first. However, even if the 
image is smoothed before computing the finite differences, the curvatures are
still sensitive to small highly curved protrusions that are unlikely to 
correspond to polyps. Thus, in this work we use a more globalized measure of
protrusion, the radius of the best fit ball. A similar \emph{sphere fitting}
approach was used in the CT colonography setting in \cite{kiss2005computer}. In
our approach we do not do the fitting to the image itself, but we first apply
a certain type of a mid-pass filter to it. This allows us to isolate the 
protrusions withing certain size limits. We use the radius of the best fit
ball as the decision parameter in a binary classifier. If the decision parameter
is larger than the discrimination threshold, then the frame is classified as
containing a polyp. 

Another feature that distinguishes our approach from the ones mentioned above 
is the use of texture information. The surface of polyps is often highly 
textured, so it makes sense to discard the frames with too little texture 
content in them. On the other hand, too much texture implies the presence of
bubbles and/or trash liquids in the frame. These unwanted features may lead
a geometry-based classifier to classify the frame as containing a polyp
when no polyp is present, i.e. they lead to an increased number of false 
positives. Thus, in order to avoid both of the situations mentioned above,
we apply a pre-selection procedure that discards the frames with too much or
too little texture content. Combined with the binary classifier this gives
the algorithm that we refer to as \emph{binary classification with pre-selection}.

\section{Details of the algorithm}
\label{sec:alg}

In the section below we present the detailed step-by-step description of the 
algorithm of processing of single frames from a capsule endoscope video 
sequence. The algorithm makes a decision for every frame whether to classify
it either as containing polyps (``polyp'' frame) or as containing normal
tissue only (``normal'' frame). 

Besides the frame to be processed, the algorithm accepts as the inputs a number 
of numerical parameters that have to be chosen in advance. The choice of these 
parameters and the robustness of the algorithm with respect to the changes in 
them is addressed in section \ref{subsec:robust}. For the purpose of numerical 
experiments, the values of these parameters were chosen manually. Ideally, we 
would like to have a systematic way to \emph{calibrate} our algorithm, i.e. to 
assign the optimal values to the parameters based on the algorithm's performance 
for some calibration data set. Currently, we do not have such a procedure, so 
this remains one of the topics of future research discussed in section 
\ref{sec:discuss}.

Another choice that requires a separate study is the choice of the color space
of the frame. In this work we convert the captured color frames to grayscale 
before processing. This choice provides good polyp detection 
results, as we observe from the numerical experiments in section \ref{sec:results}.
However, we believe that certain improvements in this area are possible. For 
example, the polyps are often highly vascularized, so one would expect them
to have a stronger red color component. Thus, one may use a measure of red 
color content in the frame, like the $a$ component of the $Lab$ color space
\cite{hunter1958photoelectric}, in polyp detection. Here we rely mostly on the
geometrical information for polyp detection, but our algorithm could still be
supplemented by the use of color information.

\subsection{Pre-processing}
\label{subsec:pre}

\begin{figure*}
\begin{flushleft}
\begin{tabular}{p{0.30\textwidth}p{0.30\textwidth}p{0.30\textwidth}p{0pt}}
\includegraphics[width=0.3\textwidth]{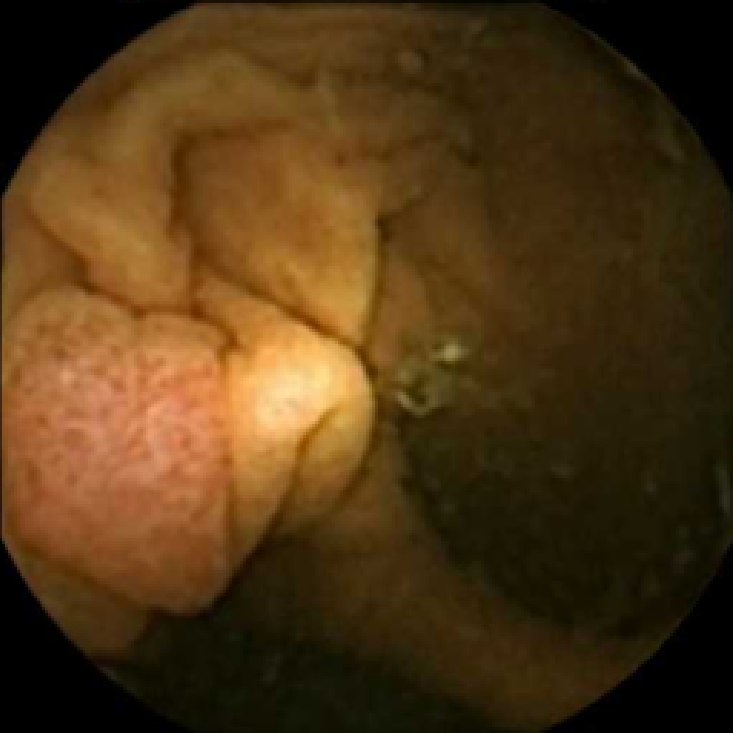} & 
\includegraphics[width=0.3\textwidth]{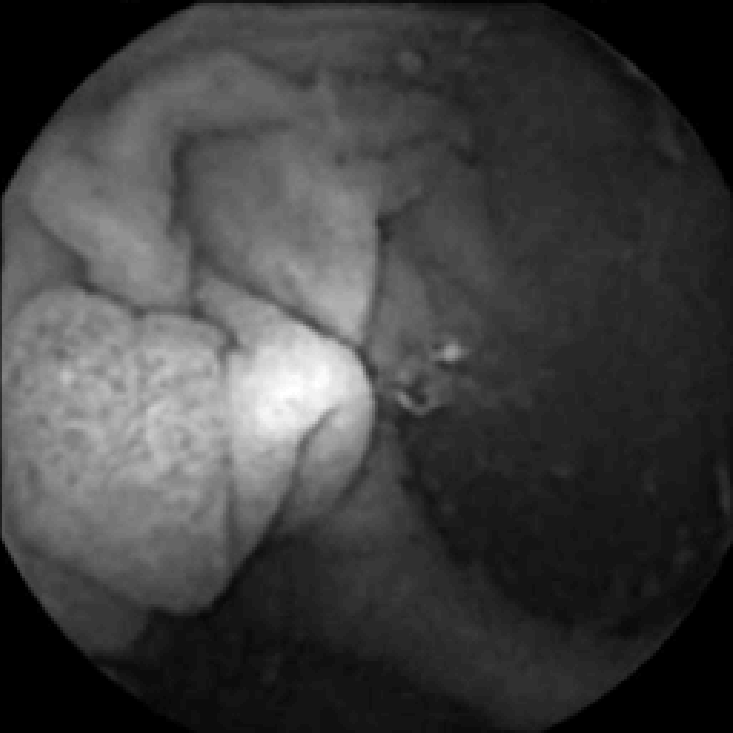} & 
\includegraphics[width=0.3\textwidth]{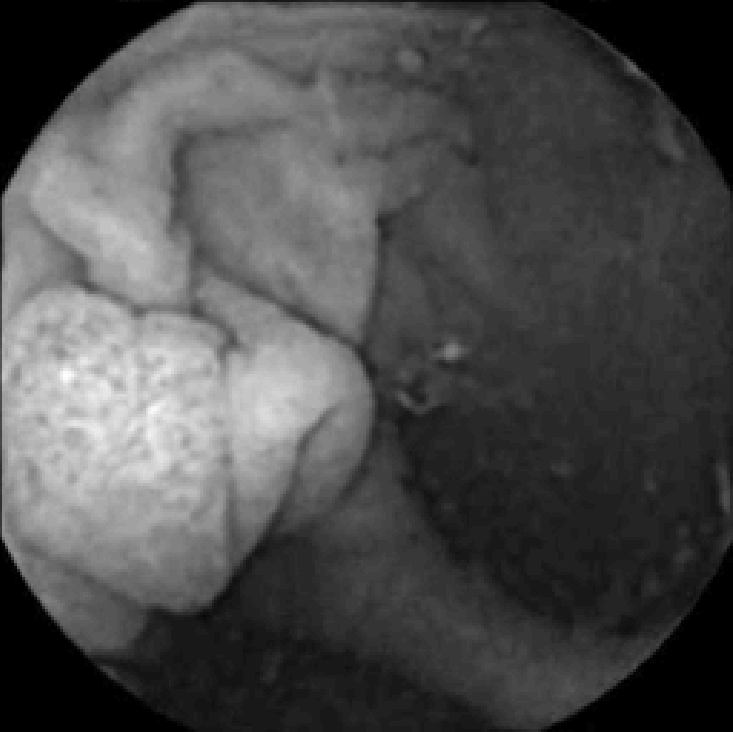} & \\
\centering (a) & \centering (b) & \centering (c) & \\
\includegraphics[width=0.3\textwidth]{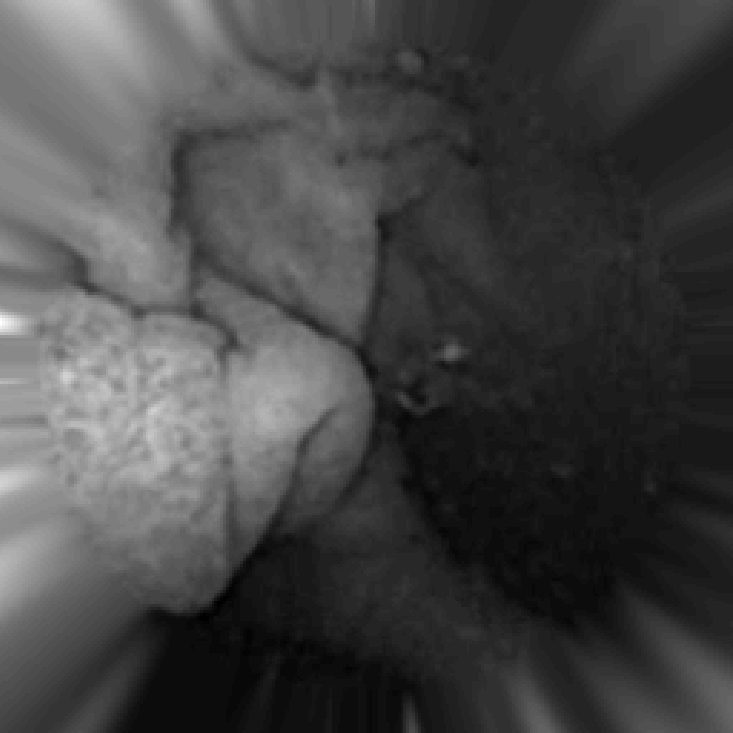} &
\includegraphics[width=0.3\textwidth]{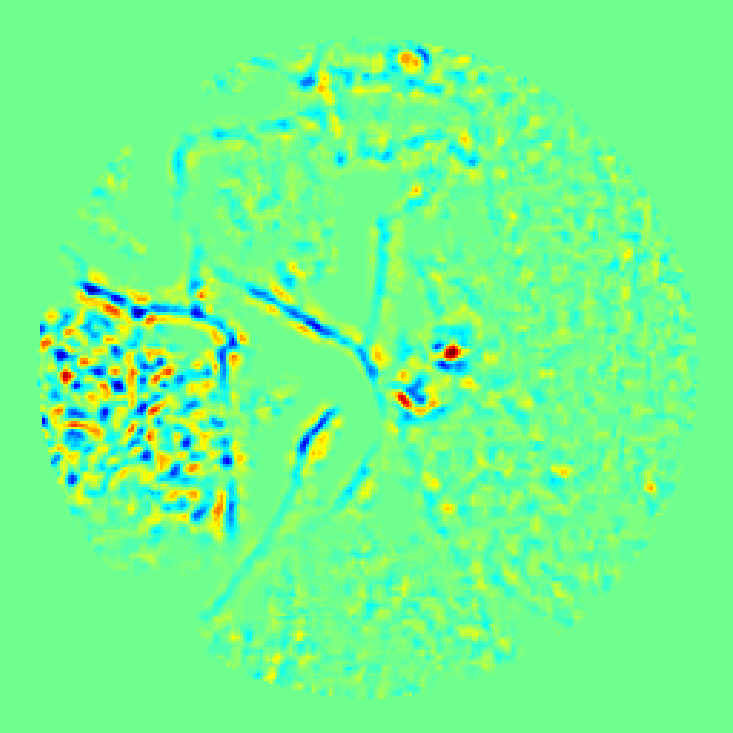} &
\includegraphics[width=0.3\textwidth]{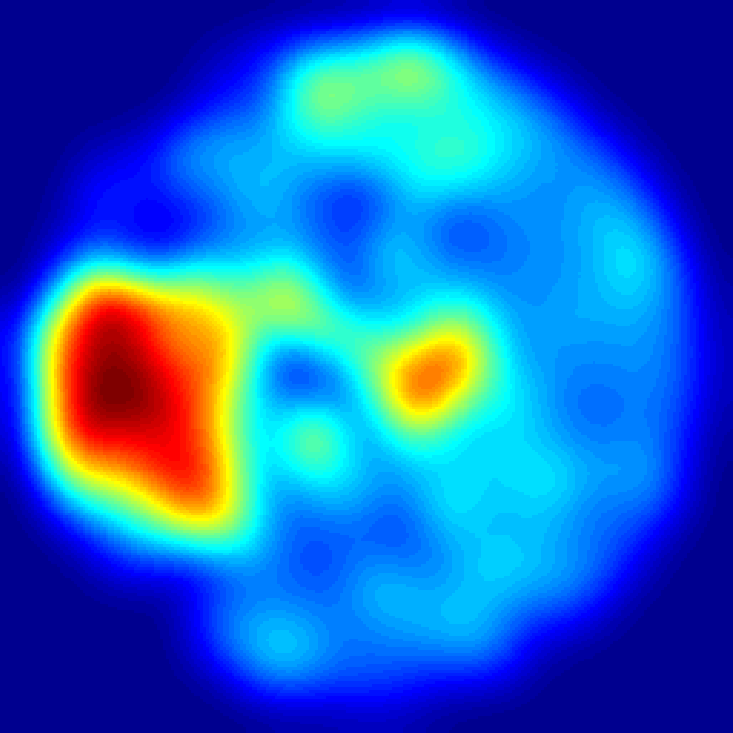} & \\
\centering (d) & \centering (e) & \centering (f) & \\
\includegraphics[width=0.3\textwidth]{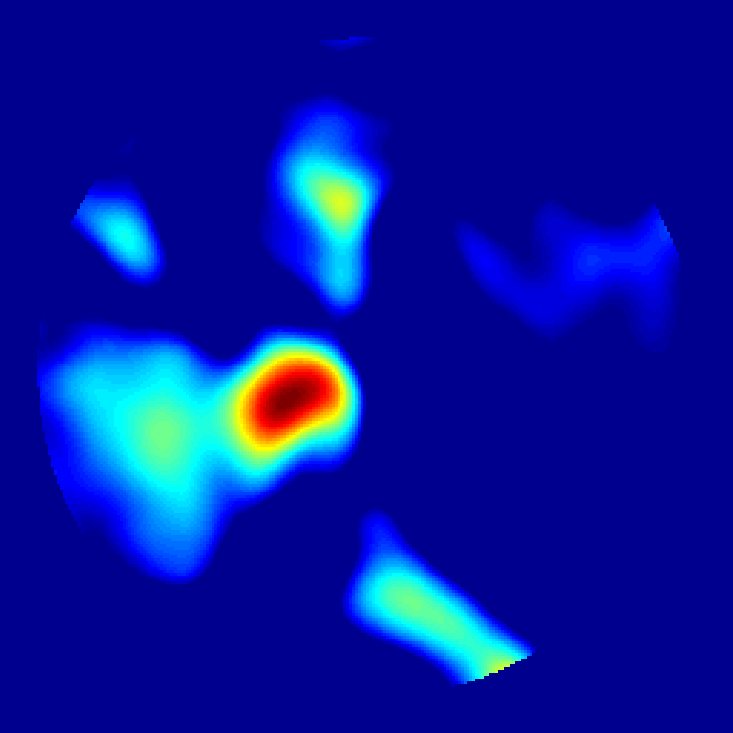} & 
\includegraphics[width=0.3\textwidth]{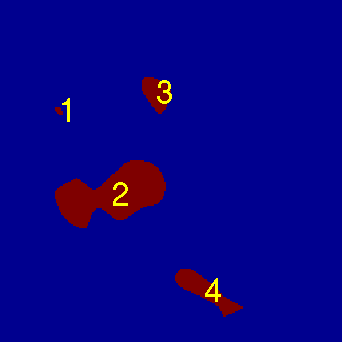} & 
\includegraphics[width=0.3\textwidth]{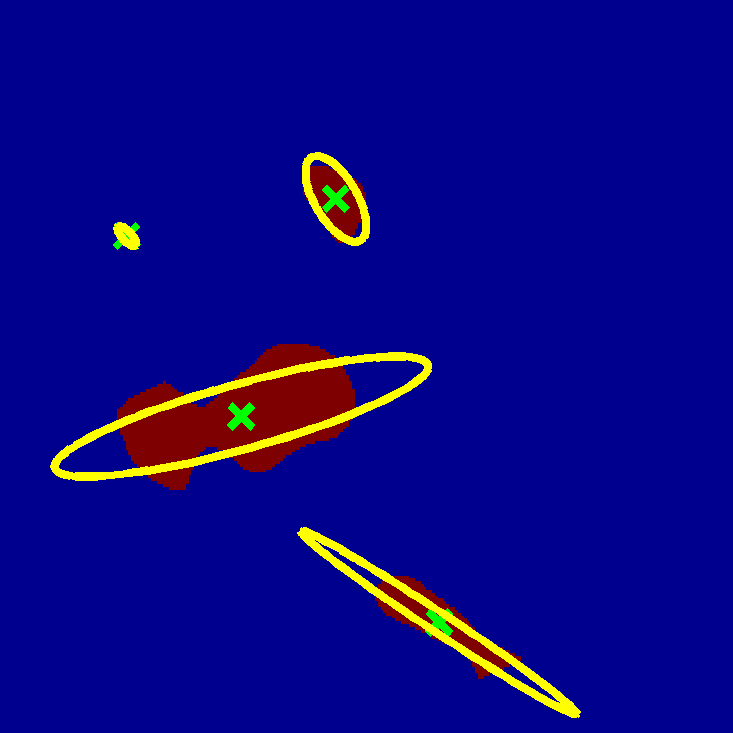} & \\
\centering (g) & \centering (h) & \centering (i) &
\end{tabular}
\end{flushleft}
\caption{Steps of the algorithm for a polyp frame: (a) original color frame, 
(b) frame in grayscale, (c) normalized intensity, 
(d) pre-processed frame $\bff$: extension by linear extrapolation, 
(e) texture $\bt$, (f) non-linear convolution-type transform $\bT$ of the texture, 
(g) mid-pass filtering $\bu$, (h) segmentation $\bs$ with $N_C=4$ connected components 
$\bs^{(k)}$ numbered $k=1,\ldots,N_C$, (i) ellipses of inertia $r^{(k)}(\theta)$ 
(yellow) with the centers of mass $(c_x^{(k)}, c_y^{(k)})$ given by green $\times$.}
\label{fig:alg-step-polyp}
\end{figure*}

Since the capsule endoscope operates in an absence of ambient light, an on-board
light source is used to capture the images. Because of the directional nature of
the light source and the optical properties of the camera's lens, the captured 
frames are often subject to an artifact known as \emph{vignetting}, which refers 
to the fall-off of intensity of the captured frame away from its center. As a 
first step of frame pre-processing we perform the normalization of intensity 
using the vignetting correction algorithm of Y. Zheng et al. 
\cite{zheng2008single}. Performance of the intensity normalization procedure 
is illustrated in Figure \ref{fig:alg-step-polyp} (c).

The images acquired by the endoscope are of circular shape. The area of the 
rectangular frame outside the circular mask is typically filled with a solid 
color. This creates a discontinuity along the edge of the circular mask, which
may cause problems in the subsequent steps of the algorithm. To remove this 
discontinuity we use a simple linear extrapolation to extend the values
from the interior of a circular mask to the rest of the rectangular frame.

This is accomplised by solving the linear system which corresponds to an upwind 
discretization of the following PDE
\begin{equation}
\nabla \bff \cdot \vec{\br} = 1,
\end{equation}
assuming that the frame $\bff$ is given on a $N_{y} \times N_{x}$ uniform Cartesian grid.
Here the vector field $\vec{\br}$ at the pixel $(i,j)$ is the unit vector 
\begin{equation}
\vec{\br}_{ij} = \frac{1}{\sqrt{(i-N_{y}/2)^{2}+(j-N_{x}/2)^{2}}} 
\begin{bmatrix} i - N_{y}/2 \\ j-N_{x}/2 \end{bmatrix}.
\end{equation}
With the values of $\bff$ inside the circular mask fixed, the solution outside 
the mask provides the desired extrapolated values. To obtain the resulting linear system 
we use a standard upwind discretization scheme; see e.g. \cite{MR1377057}.

The linear extrapolation is shown in Figure \ref{fig:alg-step-polyp} (d), where 
the radius of a circular mask $R_{mask}$ is taken to be slightly less than half 
the frame size. All the subsequent calculations involving the extrapolated 
frame are subject to masking after the calculation is done. This removes the effects 
that the artifacts of extrapolation (seen as light radial strips in the right corners 
of Figure \ref{fig:alg-step-polyp} (d)) might have on the result.

\subsection{Texture computation and convolution}
\label{subsec:texture}

Computation of the texture content in the frame is an important first step of 
the algorithm. We use the thresholding on the texture content as a pre-selection
criterion, i.e. some frames are discarded from the consideration (and labeled 
as ``normal'') based on the texture content alone.

To separate the pre-processed frame $\bff$ into the texture $\bt$ and cartoon 
$\bc$ components 
\begin{equation}
\bff = \bt + \bc,
\label{eqn:tc}
\end{equation}
we use an algorithm of Buades et al. \cite{ipol.2011.blmv_ct}.
The algorithm is based on iterative application of low-pass filtering by convolution
with a Gaussian kernel. As its input parameters it accepts the number of iterations 
$n_{iter}$ and the standard deviation $\sigma_t$ of the Gaussian kernel in pixels.
Hereafter, we treat the frame as a matrix $\bff \in \mathbb{R}^{N_y \times N_x}$,
where $N_x$ is the width and $N_y$ is the height of the frame in pixels. The
individual pixels are denoted by $f_{ij}$, $1 \leq i \leq N_y$, $1 \leq j \leq N_x$.
The quantities having the same dimensions as the frame itself ($\bt$, $\bc$, etc.) 
are treated the same way and are denoted by bold symbols. Their individual pixels
are denoted by regular symbols with indices $i$ and $j$, i.e. $t_{ij}$, $c_{ij}$, etc.

\begin{figure*}
\begin{flushleft}
\begin{tabular}{p{0.28\textwidth}p{0.31\textwidth}p{0.30\textwidth}p{0pt}}
\includegraphics[width=0.28\textwidth]{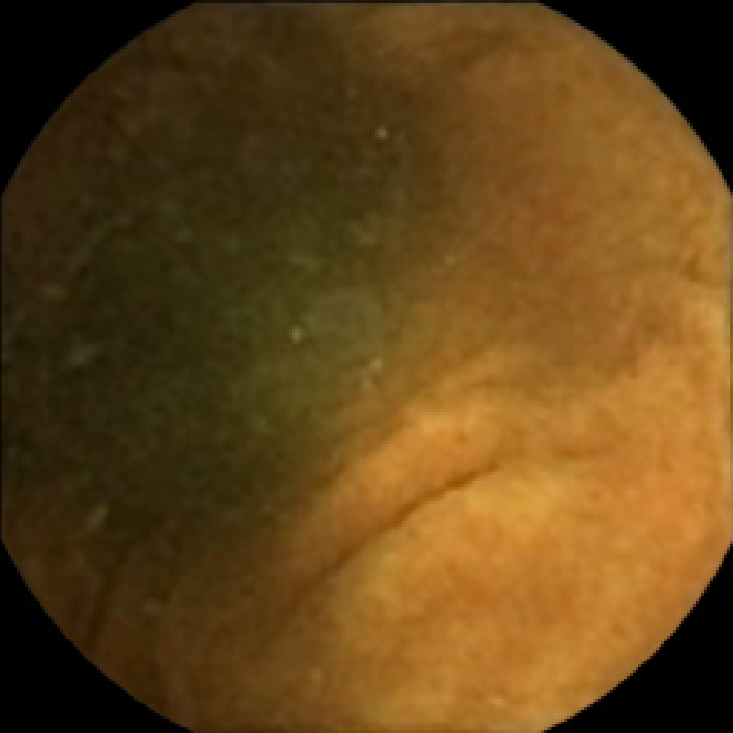} & 
\includegraphics[width=0.325\textwidth]{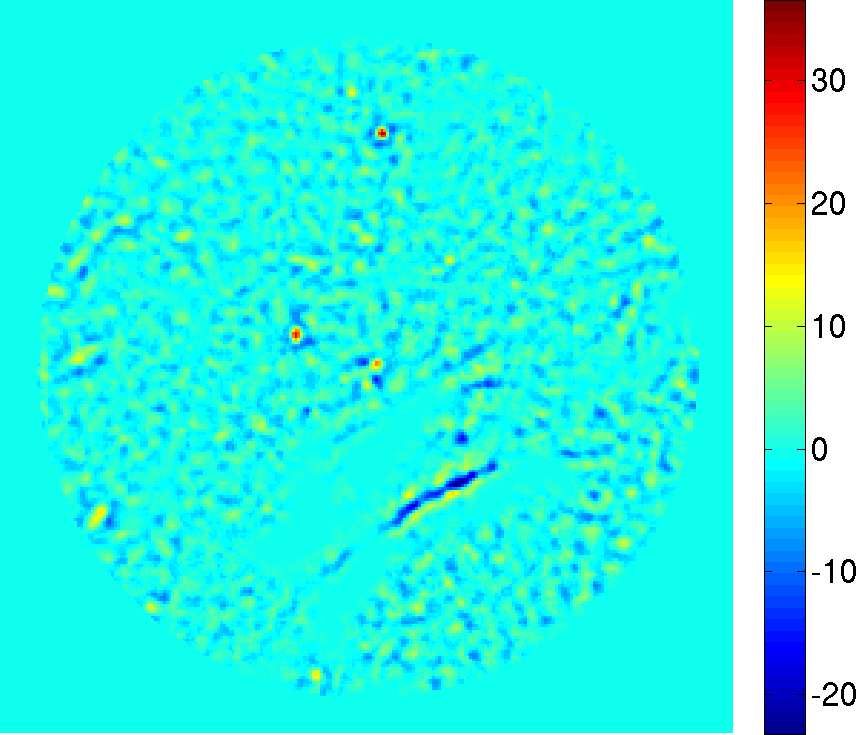} & 
\includegraphics[width=0.32\textwidth]{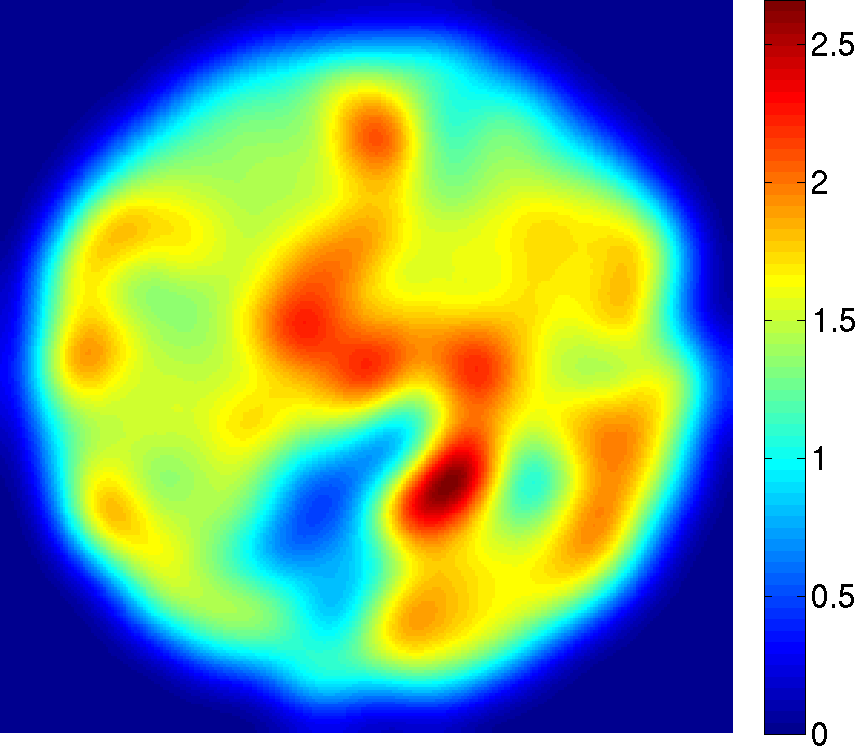} & \\
\includegraphics[width=0.28\textwidth]{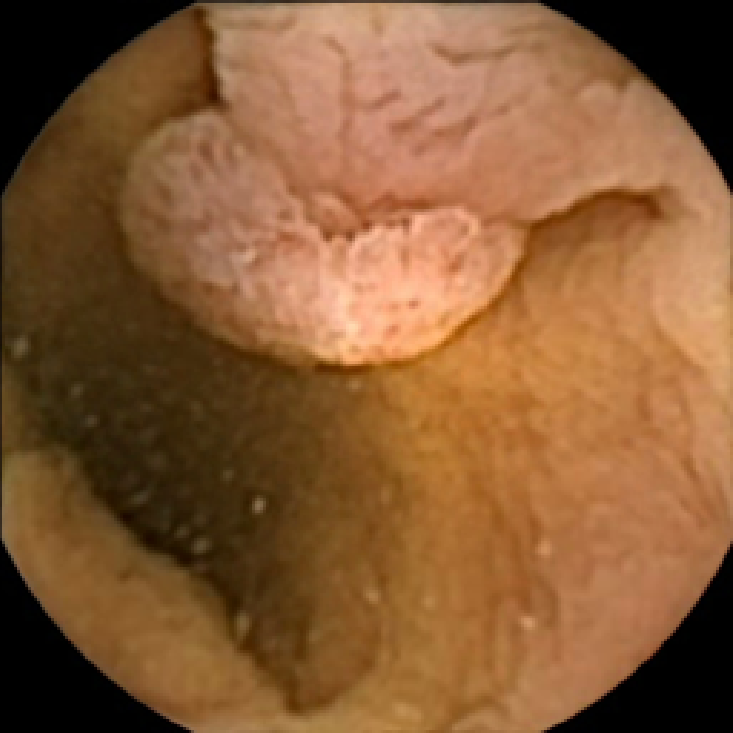} &
\includegraphics[width=0.325\textwidth]{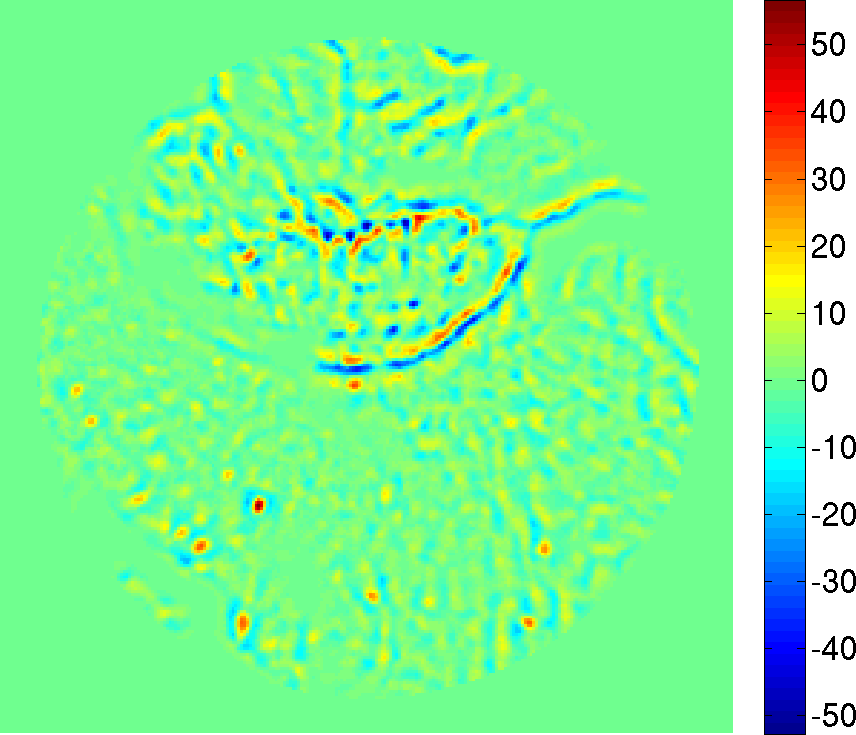} &
\includegraphics[width=0.31\textwidth]{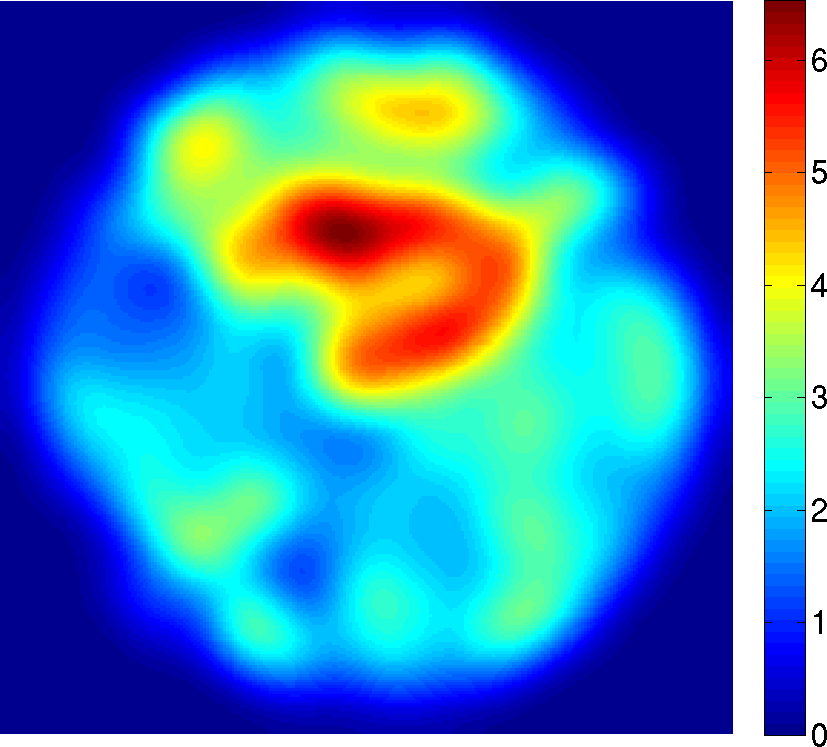} & \\
\includegraphics[width=0.28\textwidth]{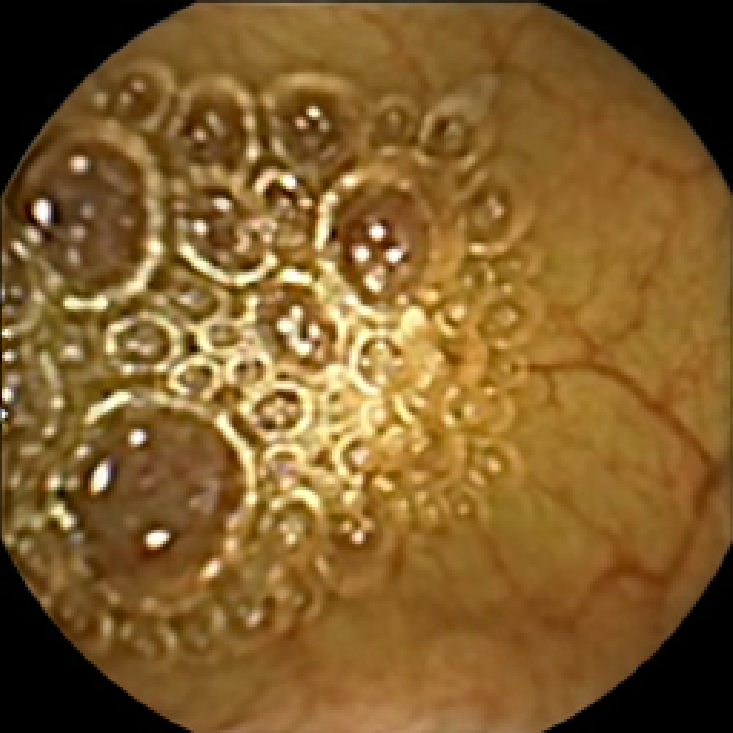} & 
\includegraphics[width=0.331\textwidth]{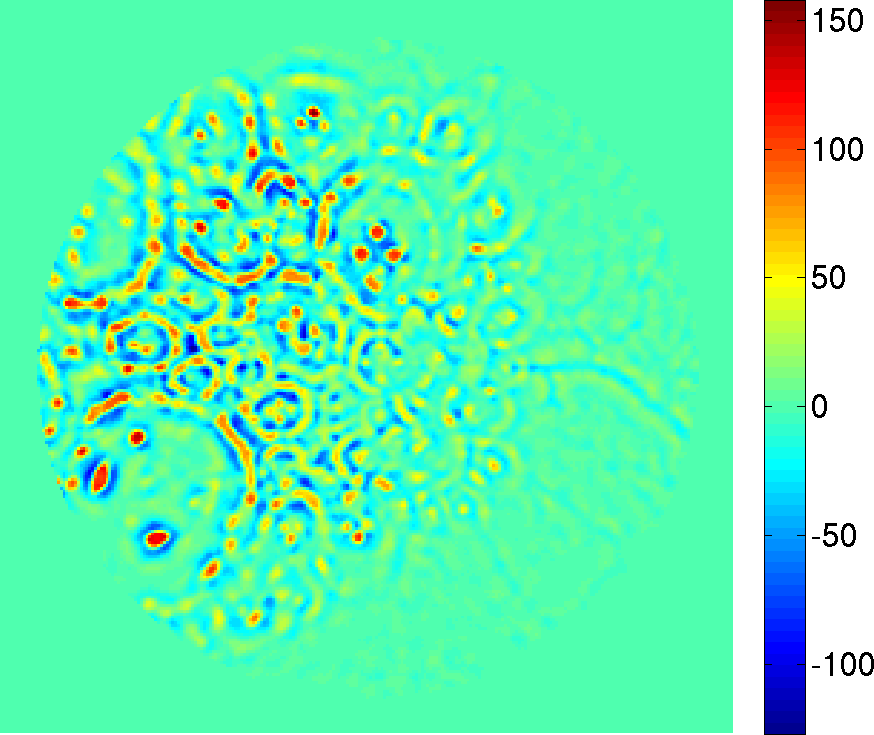} & 
\includegraphics[width=0.316\textwidth]{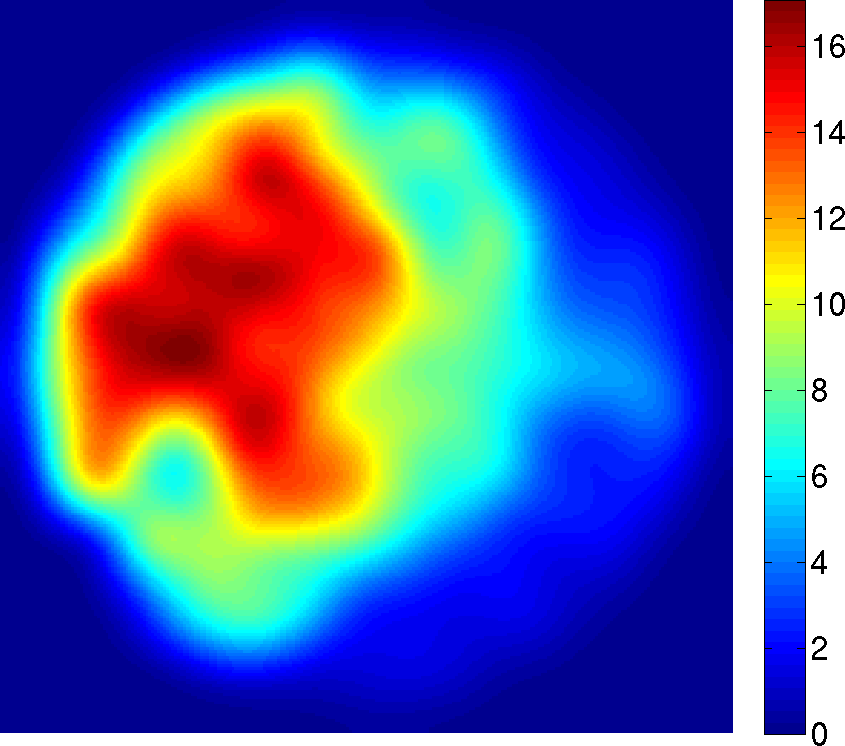} & \\
\centering (a) & \centering (b) & \centering (c) &
\end{tabular}
\end{flushleft}
\caption{Texture computation for the cases of low (top row, $T_{max}=2.6$), 
medium (middle row, $T_{max}=6.5$) and high (bottom row, $T_{max}=17.1$) texture content.
Top and bottom rows are normal frames, middle row is a polyp frame. 
Columns: (a) original frames, (b) texture $\bt$, (c) convolution-type transform $\bT$.}
\label{fig:texture}
\end{figure*}

The use of texture in pre-selection is motivated by two considerations. First,
the surface of polyps is often textured, so discarding the frames with low
texture content helps to distinguish the polyp frames from the frames with flat
mucosa. Second, when trash liquids or bubbles are present in the frame, most of
$\bff$ ends up in $\bt$, so we expect the texture content to be abnormally high in
this case. Since detecting polyps in the frames polluted with trash or bubbles
is not feasible anyway, we may as well discard the frames with very high texture
content.
Another reason to discard such frames is that the mid-pass filtering (see 
section \ref{subsec:midpass}) that we use in polyp detection is sensitive to 
the presence of large areas covered with trash and bubbles. If such frames are
not discarded, this may result in an increased number of false positives.

Once we have a decomposition (\ref{eqn:tc}), we need to define a measure of 
texture content that would be appropriate for performing the pre-selection. 
The measure should be more sensitive to the presence of large textured regions
and less sensitive to small regions even if those are strongly textured, since
those typically correspond to occasional trash liquids or bubbles. Thus, we 
perform the following non-linear convolution-type transform of the texture
\begin{equation}
\bT = L_{\sigma} (|\bt|^p),
\label{eqn:tconv}
\end{equation}
where the absolute value and exponentiation in $|\bt|^p$ are pixel-wise, and
$L_{\sigma}$ is a linear operator convolving the frame with a Gaussian kernel
with standard deviation $\sigma$. The operator $L_{\sigma}$ is also used in
mid-pass filtering in the next step of the algorithm, and is defined as follows.
First, we define a one-dimensional Gaussian kernel on a stencil of 
$2 \lceil \sigma \rceil + 1$ pixels, normalized so that it sums to one. 
Then we perform a convolution of the rows of the frame with the one-dimensional
kernel. Finally, the columns of the row-convolved frame are convolved
with the same kernel. When the stencil of the one-dimensional kernel protrudes 
outside the frame, we use mirror boundary conditions.

The usage of convolution in (\ref{eqn:tconv}) with $\sigma$ equal to a half of
a typical polyp size in pixels, allows to emphasize the textured regions that 
are likely to be polyps. 
Adding non-linearity in the form of exponentiation with $p<1$
de-emphasizes small regions with strong texture. This is illustrated in 
Figure \ref{fig:alg-step-polyp} (e) and (f), where $\bt$ and $\bT$ are shown 
respectively. We observe that there are two well-pronounced bumps in $\bT$. A 
larger in size and magnitude on the left corresponds to the polyp, a smaller
one in the middle-right is due to a few bubbles present in the frame.

\begin{figure}
\centering
\includegraphics[width=0.45\textwidth]{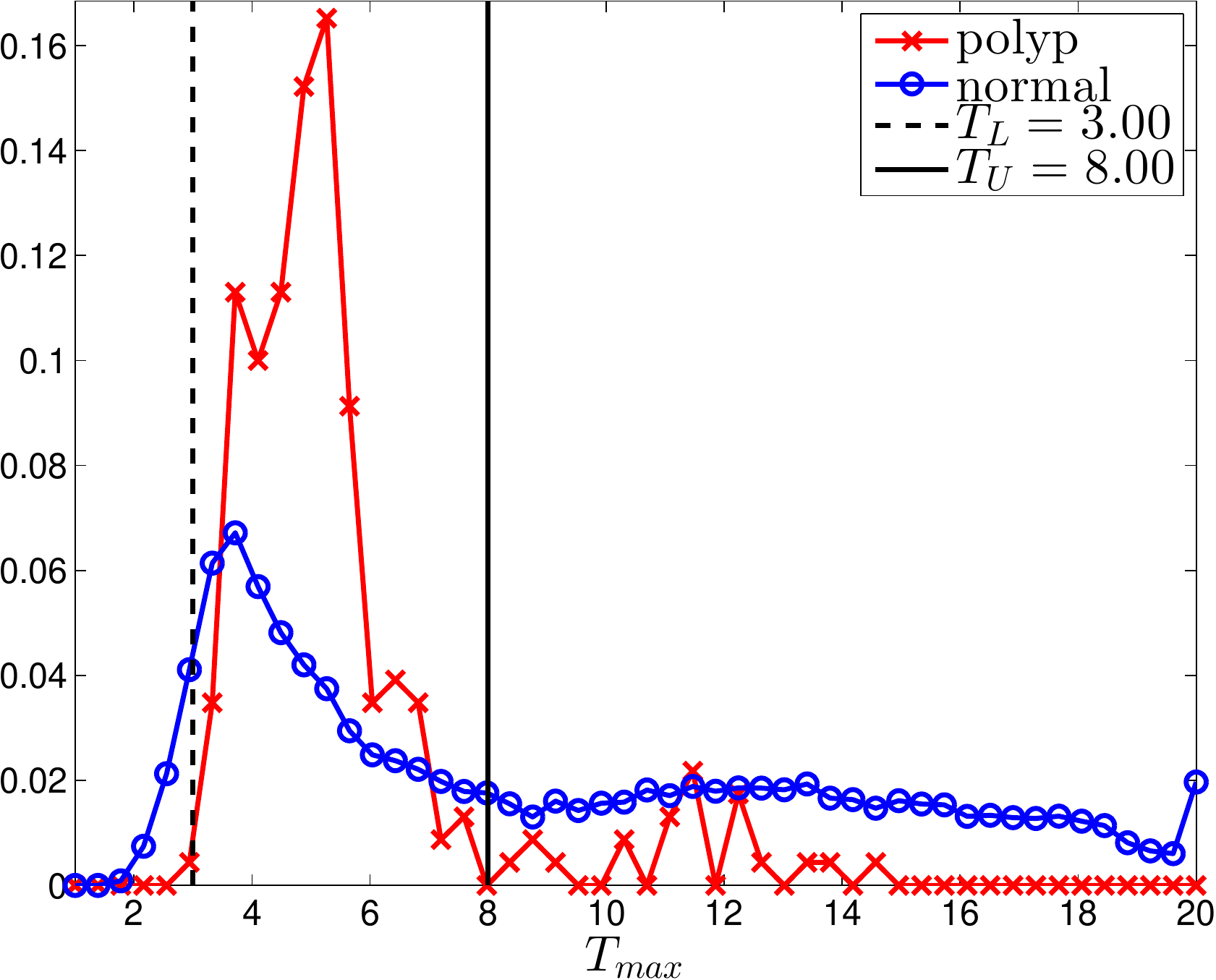}
\caption{Histogram of the distribution of $T_{max}$ for normal and polyp frames.}
\label{fig:hist-texture}
\end{figure}

Once the non-linear transform (\ref{eqn:tconv}) is calculated, we can compute
the measure of texture content
\begin{equation}
T_{max} = \mathop{\mbox{max }}\limits_{i,j} T_{ij}, 
\quad 1\leq i \leq N_y,\; 1 \leq j \leq N_x.
\label{eqn:tmax}
\end{equation}
Then the \emph{pre-selection criterion} is a simple thresholding
\begin{equation}
T_L \leq T_{max} \leq T_U.
\label{eqn:tcriterion}
\end{equation}
The lower bound filters out the frames with too little texture content that are
unlikely to contain any polyps due to most polyps having a textured surface.
The upper bound allows us to discard the frames polluted with trash and bubbles, 
since even if they contain polyps, they are likely to be obscured. This is illustrated 
in Figure \ref{fig:texture}, where we display two normal frames with low and high 
values of $T_{max}$ and a polyp frame with a medium value of $T_{max}$. As expected, 
the first normal frame containing flat mucosa has little texture content. The second
normal frame polluted with bubbles has strong texture content in the bubbles area,
which is especially pronounced in the plot of $\bT$. Finally, the polyp frame has 
moderately textured polyp area, which can also be easily observed from $\bT$ that has
the strongest feature in that region.

To verify the above conclusions statistically, we compare in Figure 
\ref{fig:hist-texture} the histograms of the distributions of $T_{max}$ for 
normal and polyp frames in our test data set  (see section \ref{subsec:dataset} 
for a detailed description). We observe that 
the peak of the histogram for normal frames is shifted towards the lower values of 
$T_{max}$, which explains the effectiveness of the lower threshold. For high
values of $T_{max}$ the histogram for normal frames is consistently well above
the one for the polyp frames, indicating a large number of frames polluted with
trash and bubbles. Given such distributions of $T_{max}$, the pre-selection 
criterion (\ref{eqn:tcriterion}) appears quite effective. For the values of 
parameters given in Table \ref{tab:params}, exactly $90\%$ of the polyp frames
pass the pre-selection, while only $47.84\%$ of the normal frames do so.

\subsection{Mid-pass filtering and segmentation}
\label{subsec:midpass}

After the frame passes the pre-selection, we identify certain regions that may
correspond to polyps. An essential feature of polyps is that they are 
protrusions or bumps on a flatter surrounding tissue. The purpose of this step 
is to detect such geometric features. Note that the polyps have a certain range 
of characteristic dimensions. Thus, in order to detect possible polyps, the 
geometrical processing should act as a \emph{mid-pass filter} that filters out
the features that are too small or too large. Here we use a mid-pass filter of 
the form
\begin{equation}
\bu = H(\bw) \cdot \bw,
\label{eqn:midpass}
\end{equation}
where $\bw$ is defined by 
\begin{equation}
\bw = \frac{L_{\sigma_1} (\bff)}{L_{\sigma_2} (\bff)} - 1,
\label{eqn:gaussratio}
\end{equation}
and $H$ is the Heaviside step function
\begin{equation}
H(x) = \left\{\begin{tabular}{ll}
0, & if $x < 0$,\\
1, & if $x \geq 0$.
\end{tabular}
\right.
\end{equation}
The application of $H$, multiplication and division in (\ref{eqn:midpass}) are 
pixel-wise. The standard deviations of the convolution
operators satisfy $\sigma_1 < \sigma_2$. They correspond to the typical radii 
(in pixels) of the polyps that we expect to detect. 
Note that measuring the polyp size in pixels is sensitive to the distance 
between the polyp and the camera, especially when such distance is large. 
This may lead to failure to detect the polyps which size in pixels is small.
Here we rely on the fact that the polyp is likely to be present in a 
number of consecutive frames as the capsule moves along. Thus, the polyp will 
be observed from different distances including some that are small enough to
make the pixel size sufficient for producing a detectable feature in the 
mid-pass filtered frame $\bu$.

We use a ratio in (\ref{eqn:midpass}) instead of a difference to obtain a 
quantity that depends less on the absolute prominence of the protrusion, but 
more on its relative prominence compared to the surrounding tissue. This allows 
for a better detection of flat polyps. Also, the difference of Gaussians is 
known to be subject to the scaling effect \cite{lindeberg1993scale} similarly 
to the Laplacian of a Gaussian. However, the ratio in (\ref{eqn:gaussratio}) is 
invariant under the scaling of the image $\bff$, thus the scaling effect does 
not apply in our case. 

Since the convolution with a smaller standard deviation is in the numerator 
of (\ref{eqn:gaussratio}), the protrusions correspond to large positive values
of $\bw$, hence we only use of the positive part of $\bw$ in (\ref{eqn:midpass}).

In Figure \ref{fig:alg-step-polyp} (g) we show the result of mid-pass
filtering $\bu$. We observe several features present in $\bu$ with the most 
prominent one corresponding to a polyp. To perform the binary classification
we need to assign a numerical quantity to each of these features that would 
determine how likely does each of them correspond to a polyp. 

To separate the features from each other we use a \emph{binary segmentation} 
via thresholding
\begin{equation}
\bs = H(\bu - \Theta) \in \{0,1\}^{N_y \times N_x}, 
\label{eqn:segment}
\end{equation}
where $H$ is taken pixel-wise and the scalar threshold $\Theta$ is defined by
\begin{equation}
\begin{split}
\Theta = & \mbox{max} \left( \mbox{min} \left( \frac{1}{2} 
\mathop{\mbox{max}}\limits_{i,j} u_{ij}, M_U \right), M_L \right),\\
& 1\leq i \leq N_y,\; 1 \leq j \leq N_x,
\end{split}
\label{eqn:theta}
\end{equation}
with some bounds $M_U > M_L > 0$. This means that the threshold $\Theta$ is
taken to be a half of maximum value of $\bu$, unless it goes above $M_U$ or 
below $M_L$, in which case it defaults to the corresponding bound. 

An example of a binary segmentation $\bs$ obtained with (\ref{eqn:segment}) is 
shown in Figure \ref{fig:alg-step-polyp} (h), where four features can be seen. 
By features we mean the connected components of $\bs$, which can be found using an 
algorithm by Haralick and Shapiro \cite{haralick1992lg}. It provides a 
decomposition
\begin{equation}
\bs = \sum_{k=1}^{N_C} \bs^{(k)},
\label{eqn:conncomp}
\end{equation}
where $N_C$ is the total number of connected components in $\bs$, and $\bs^{(k)}$
are the disjoint connected components. The pixel values $s_{ij}^{(k)}$ of 
$\bs^{(k)}$ are defined as
\begin{equation}
s_{ij}^{(k)} = \left\{
\begin{tabular}{ll}
1, & \begin{tabular}{l} if pixel $(i,j)$ belongs to the\\$k^{th}$ connected component, \end{tabular} \\
0, & \begin{tabular}{l} otherwise. \end{tabular}
\end{tabular}
\right.
\end{equation}
Decomposition (\ref{eqn:conncomp})
is illustrated in Figure \ref{fig:alg-step-polyp} (h), where the four features 
$\bs^{(k)}$ are numbered $k=1,\ldots,N_C$ in the order they are found by the 
algorithm.

\subsection{Geometrical processing and the tensor of intertia}
\label{subsec:geomproc}

After the binary segmentation of the frame is decomposed into separate features
(\ref{eqn:conncomp}), we can process them individually to determine which of 
them might correspond to polyps. The simplest criterion we can apply is 
filtering the features by their sizes
\begin{equation}
K_S = \left\{ k \in \{1,2,\ldots,N_C\} \;|\; S_L \leq S^{(k)} \leq S_U \right\},
\label{eqn:size-criterion}
\end{equation}
where the size $S^{(k)}$ of the $k^{th}$ feature is defined by
\begin{equation}
S^{(k)} = \sum_{i,j} s_{ij}^{(k)}, \quad k = 1,\ldots,N_C.
\label{eqn:size}
\end{equation}

\begin{figure*}
\begin{flushleft}
\begin{tabular}{p{0.22\textwidth}p{0.22\textwidth}p{0.22\textwidth}p{0.22\textwidth}p{0pt}}
\includegraphics[width=0.23\textwidth]{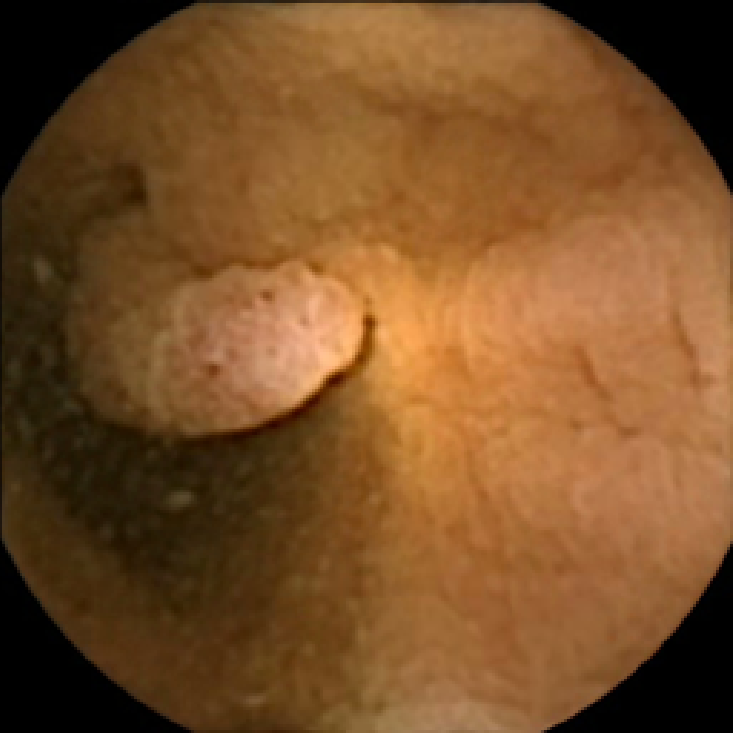} & 
\includegraphics[width=0.23\textwidth]{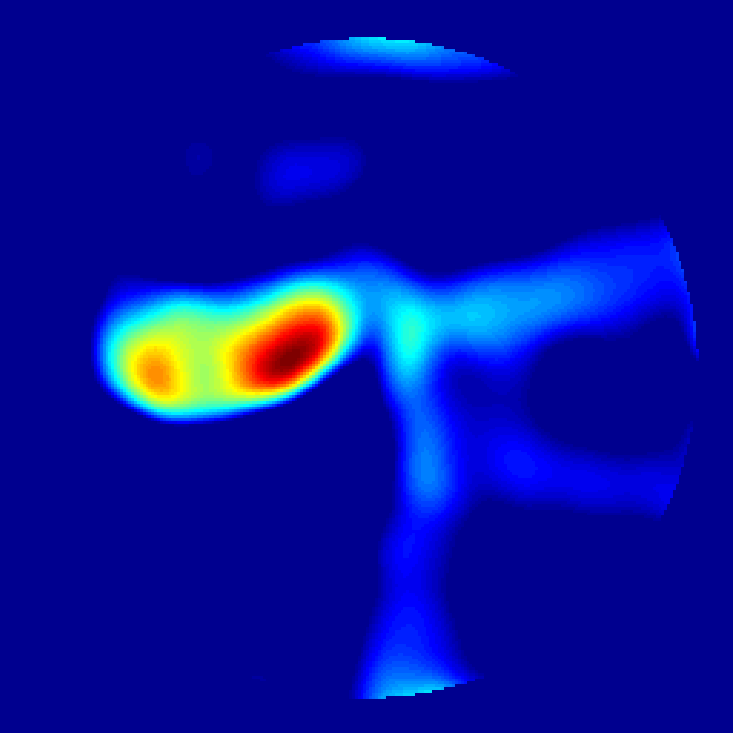} & 
\includegraphics[width=0.23\textwidth]{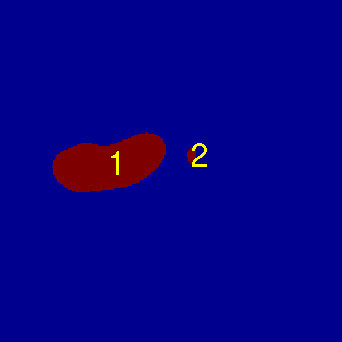} &
\includegraphics[width=0.23\textwidth]{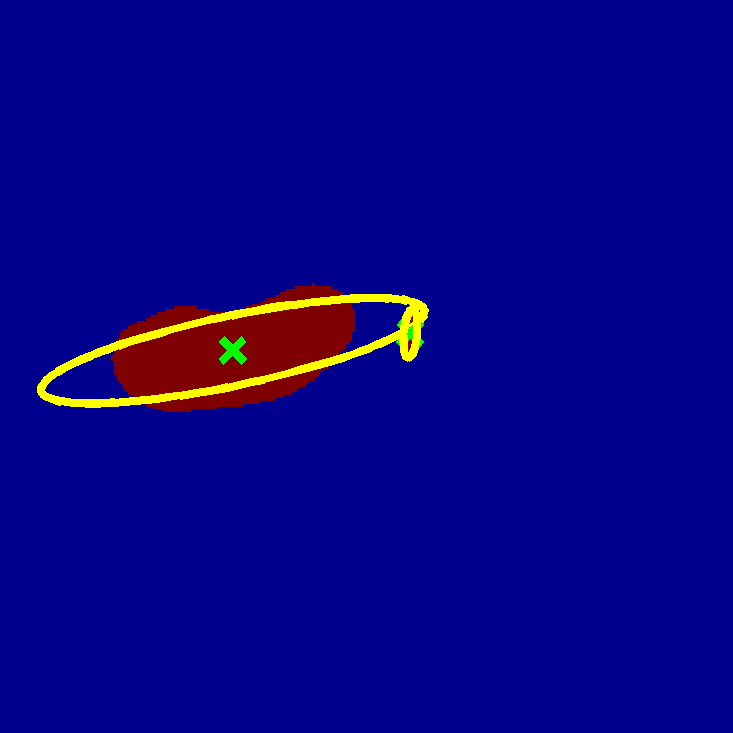} & \\
\includegraphics[width=0.23\textwidth]{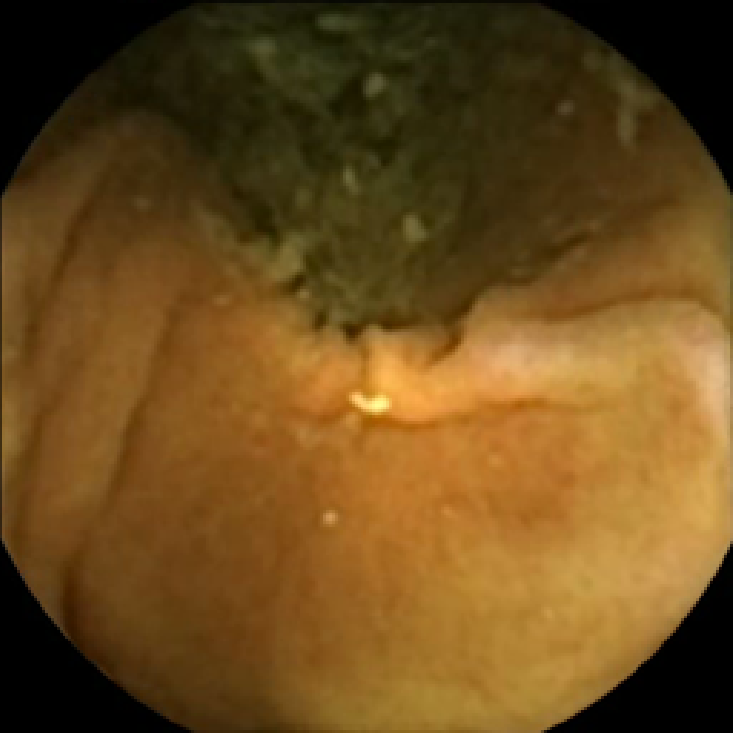} &
\includegraphics[width=0.23\textwidth]{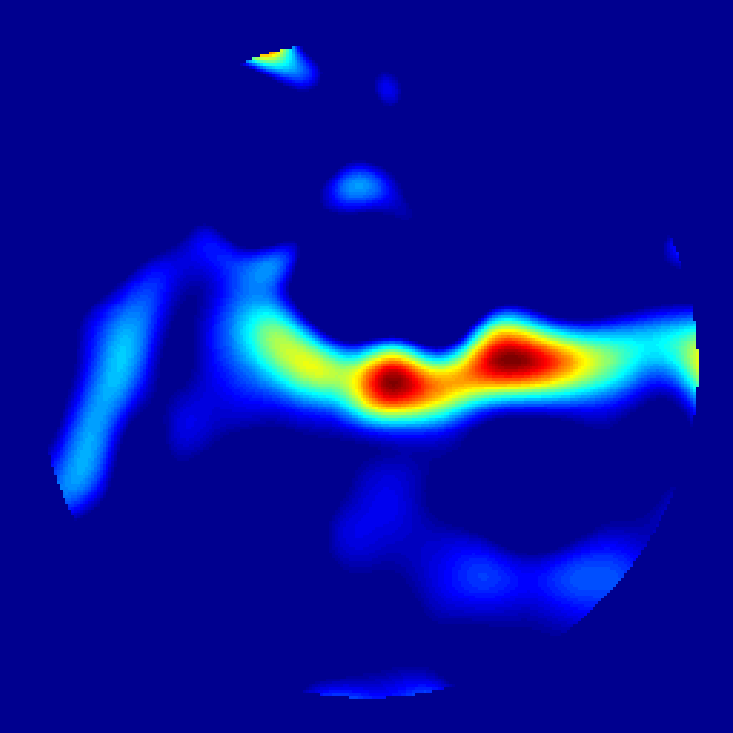} &
\includegraphics[width=0.23\textwidth]{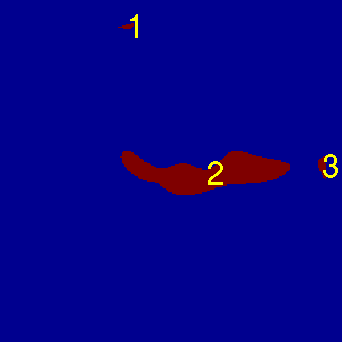} & 
\includegraphics[width=0.23\textwidth]{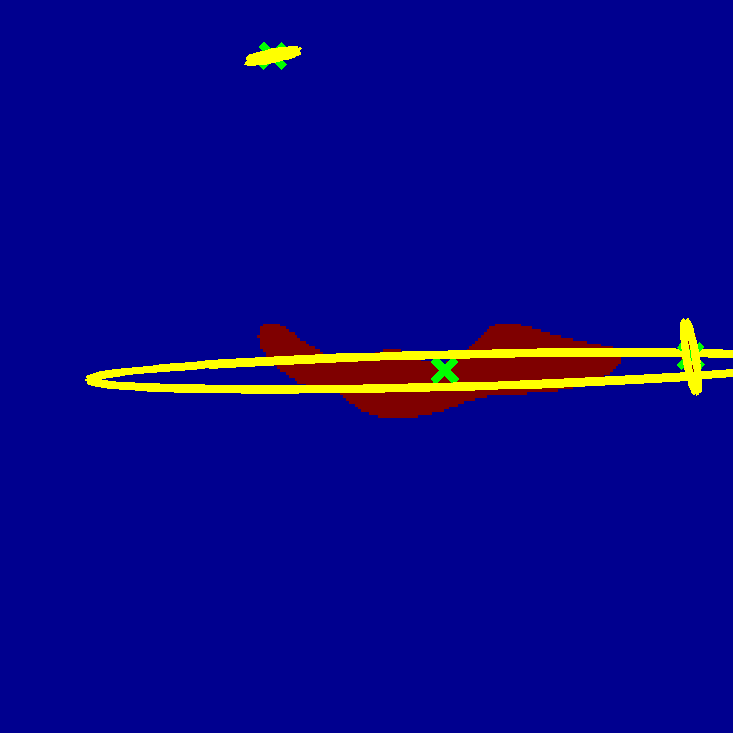} & \\
\includegraphics[width=0.23\textwidth]{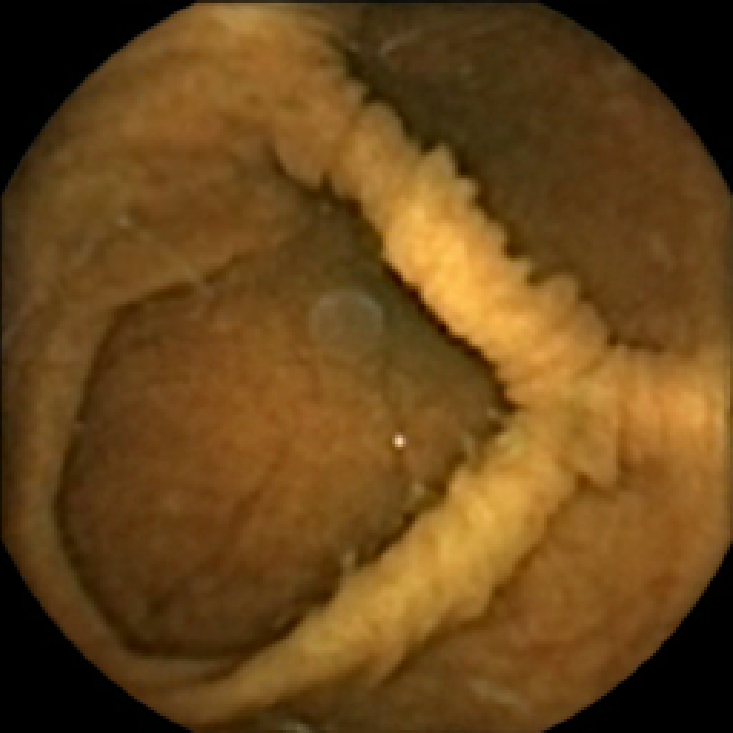} &
\includegraphics[width=0.23\textwidth]{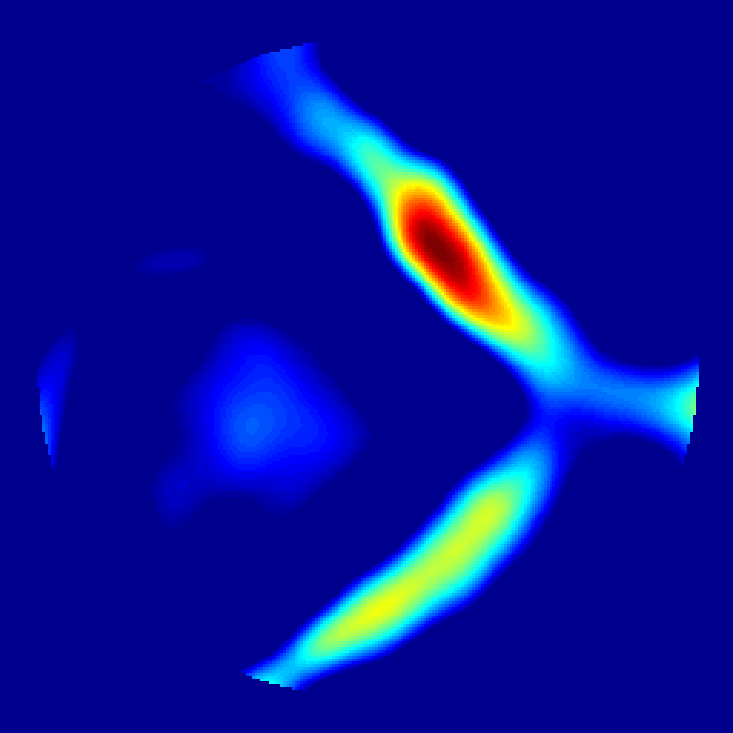} &
\includegraphics[width=0.23\textwidth]{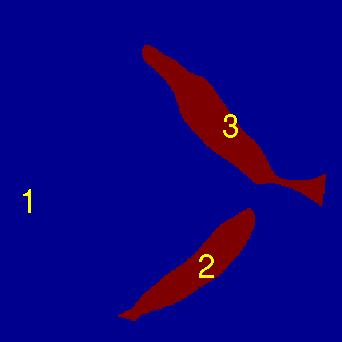} & 
\includegraphics[width=0.23\textwidth]{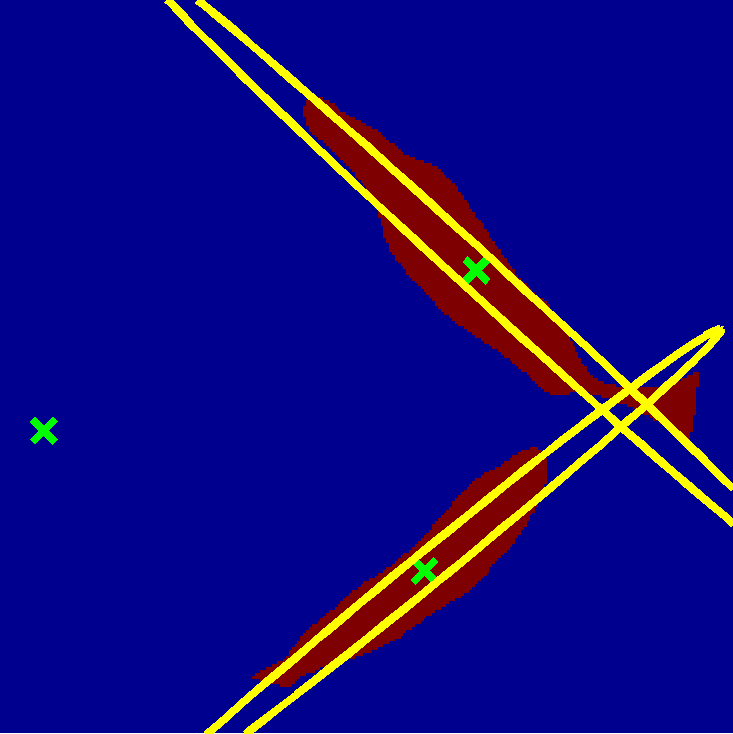} & \\
\centering (a) & \centering (b) & \centering (c) & \centering (d) & 
\end{tabular}
\end{flushleft}
\caption{Comparison of the geometrical processing and the tensor of inertia 
calculation for a polyp frame (first row) and two normal frames (second and third rows). 
Columns: (a) original frames, (b) mid-pass filtering $\bu$, (c) binary segmentation 
$\bs = \sum_{k=1}^{N_C} s^{(k)}$, (d) ellipses of inertia $r^{(k)}(\theta)$ (yellow)
with the centers of mass $(c_x^{(k)}, c_y^{(k)})$ given by green $\times$.}
\label{fig:geom}
\end{figure*}

Features that are too large, typically correspond to folds of normal mucosal 
tissue. Very small features are likely to be the artifacts of mid-pass filtering
and the subsequent segmentation. The above feature size criterion can be used 
to discard such features. 

A more sophisticated criterion that can help to eliminate the non-polyp features
that pass the size criterion (\ref{eqn:size-criterion}) is based on the 
computation of the features' \emph{tensors of inertia} \cite{pauly2003point} 
to determine how stretched the feature is. For this we define the matrices
\begin{equation}
\begin{split}
x^{(k)}_{ij} = & \left\{
\begin{tabular}{ll}
$j$, & if $s^{(k)}_{ij} = 1$ \\
$0$, & if $s^{(k)}_{ij} = 0$
\end{tabular}\right. ,\\
y^{(k)}_{ij} = & \left\{
\begin{tabular}{ll}
$i$, & if $s^{(k)}_{ij} = 1$ \\
$0$, & if $s^{(k)}_{ij} = 0$
\end{tabular}\right. ,\\
& 1 \leq i \leq N_y, \; 1 \leq j \leq N_x,
\end{split}
\label{eqn:xy}
\end{equation}
that allow us to compute first the \emph{centers of mass}
\begin{equation}
\begin{split}
c_x^{(k)} = & \frac{1}{S^{(k)}} \sum\limits_{i,j} x^{(k)}_{ij}, \\
c_y^{(k)} = & \frac{1}{S^{(k)}} \sum\limits_{i,j} y^{(k)}_{ij}, \\
& k = 1,\ldots,N_C.
\end{split}
\label{eqn:cmass}
\end{equation}

Then we can define the tensors of inertia $I^{(k)} \in \mathbb{R}^{2 \times 2}$ as
\begin{equation}
\begin{split}
I^{(k)} = & \sum_{i,j} \begin{bmatrix} 
\left( \widehat{y}^{(k)}_{ij} \right)^2 & 
- \widehat{x}^{(k)}_{ij} \widehat{y}^{(k)}_{ij} \\
- \widehat{x}^{(k)}_{ij} \widehat{y}^{(k)}_{ij} &
\left( \widehat{x}^{(k)}_{ij} \right)^2
\end{bmatrix}, \\ 
& k = 1,\ldots,N_C,
\end{split}
\label{eqn:tensor}
\end{equation}
where $\widehat{x}^{(k)}_{ij} = x^{(k)}_{ij} - c_x^{(k)}$ and 
$\widehat{y}^{(k)}_{ij} = y^{(k)}_{ij} - c_y^{(k)}$ are the coordinates relative
to the centers of mass. The tensors of inertia are
symmetric positive definite, thus they can be used to define ellipses. Using an
analogy from the classical mechanics, we refer to these ellipses as the 
\emph{ellipses of inertia}. The eccentricities $E^{(k)}$ of such ellipses 
are given by
\begin{equation}
E^{(k)} = \frac{\lambda^{(k)}_{max}}{\lambda^{(k)}_{min}}, \quad k = 1,\ldots,N_C,
\label{eqn:ecc}
\end{equation}
where $\lambda^{(k)}_{max} \geq \lambda^{(k)}_{min} > 0$ are the eigenvalues of
$I^{(k)}$. The eccentricity $E^{(k)}$ determines how much is the $k^{th}$ feature
stretched in one direction compared to its transversal. This information is useful
for our purposes, since we expect the polyps to be more round in shape than the
mucosal folds, that are often stretched.

We illustrate the above considerations in Figure \ref{fig:geom}, where we compare
the ellipses of inertia for a polyp frame and two frames with pronounced mucosal
folds. The ellipses we plot are
\begin{equation} 
r^{(k)} (\theta) =
\sqrt{ \frac{S^{(k)}}{\pi \lambda_{max}^{(k)} \lambda_{min}^{(k)}} } \; 
I^{(k)} \begin{pmatrix} \cos \theta \\ \sin \theta \end{pmatrix} +
\begin{pmatrix} c^{(k)}_x \\ c^{(k)}_y \end{pmatrix},
\end{equation}  
where $\theta \in [0, 2\pi]$. The scaling term in front of $I^{(k)}$ is chosen 
so that the area of the ellipse of inertia is the same as the size $S^{(k)}$ of the 
corresponding feature.

As expected, we observe that the ellipses corresponding to mucosal folds 
(feature 2 in the second row and features 2 and 3 in the third row of Figure 
\ref{fig:geom}) are indeed much more stretched out than the ellipse corresponding 
to a polyp (feature 1 in the first row of Figure \ref{fig:geom}). Stretched ellipses 
imply higher eccentricity, thus we impose the following criterion
\begin{equation}
K_E = \left\{ k \in \{1,2,\ldots,N_C\} \;|\; E^{(k)} \leq E_{max} \right\}
\label{eqn:ecc-criterion}
\end{equation}
with some threshold $E_{max}$ to select moderately stretched features that 
are more likely to correspond to polyps.

The combined geometric criterion is
\begin{equation}
K_G = K_S \cap K_E.
\label{eqn:geom-criterion}
\end{equation}
If none of the features in the frame passes this criterion, i.e. if 
$K_G = \varnothing$, then the frame is labeled as normal. If one or more features 
satisfy (\ref{eqn:geom-criterion}), then we continue to the next step, where 
we compute a parameter upon which we base the decision whether the frame
is classified as containing polyps or not.

\subsection{Decision parameter and binary classifier}
\label{subsec:ball}

The final step of the algorithm is the computation of the decision parameter 
that we use in the binary classification. This parameter is geometrical in nature 
and we define it as follows. 

\begin{figure*}
\begin{flushleft}
\begin{tabular}{p{0.22\textwidth}p{0.22\textwidth}p{0.22\textwidth}p{0.22\textwidth}p{0pt}}
\includegraphics[width=0.23\textwidth]{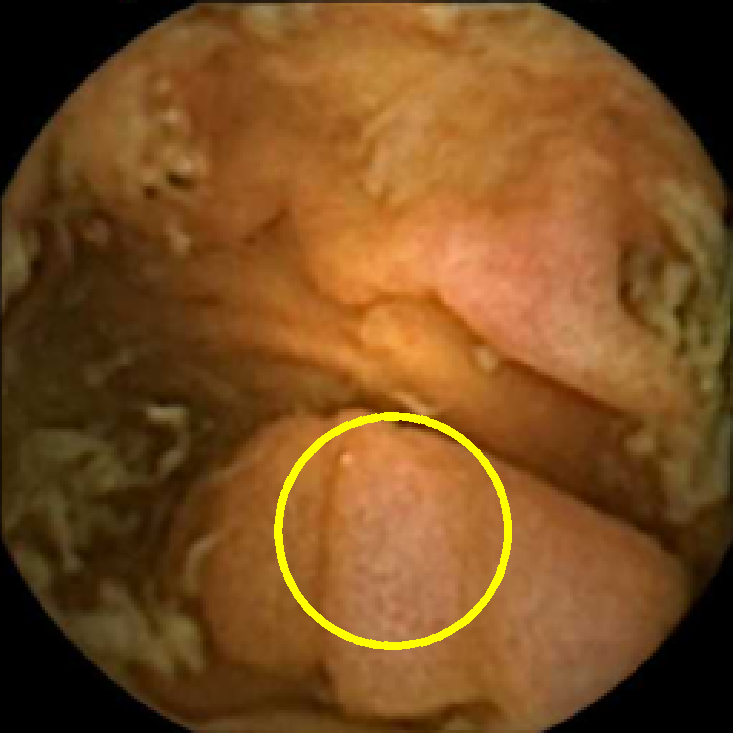}  & 
\includegraphics[width=0.23\textwidth]{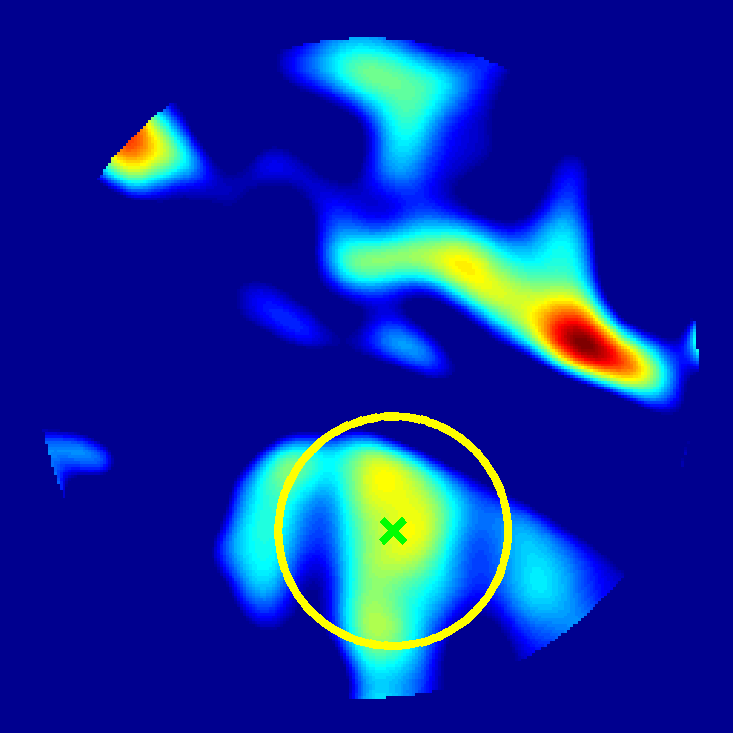}  & 
\includegraphics[width=0.23\textwidth]{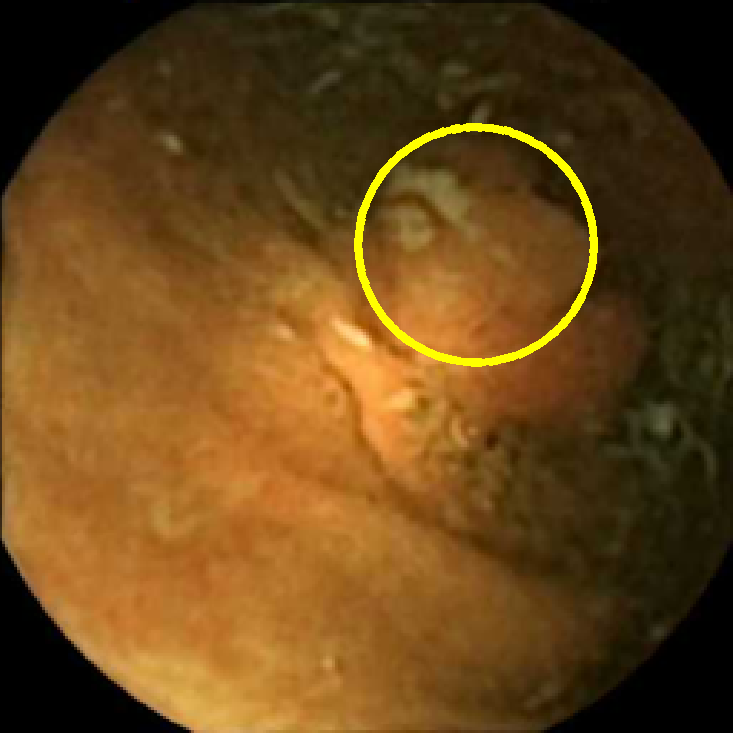}  &
\includegraphics[width=0.23\textwidth]{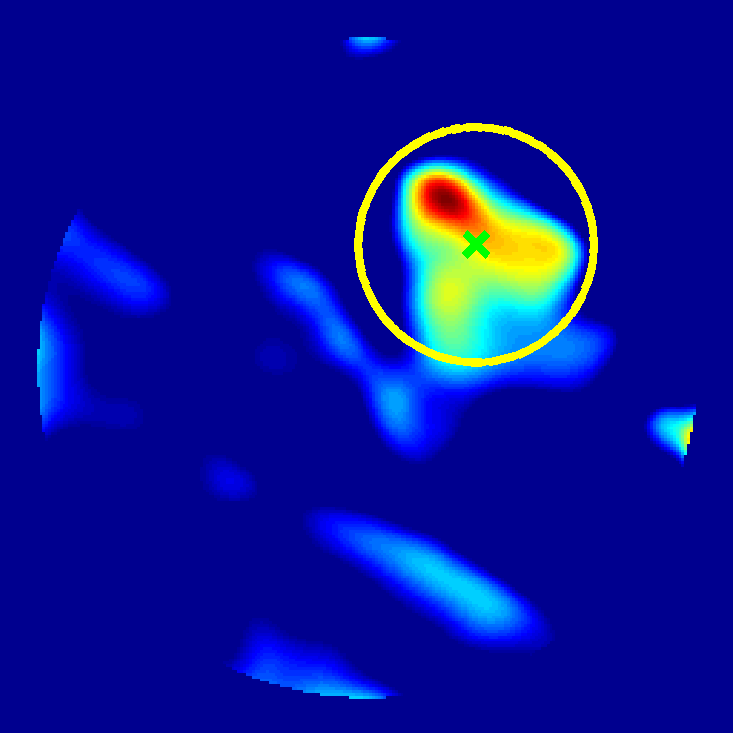}  & \\  
\includegraphics[width=0.23\textwidth]{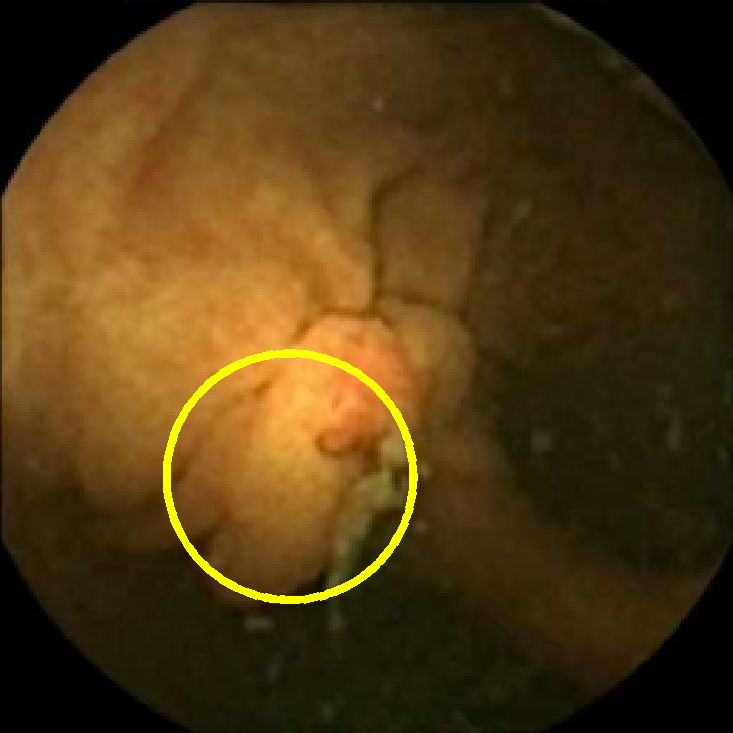}  &
\includegraphics[width=0.23\textwidth]{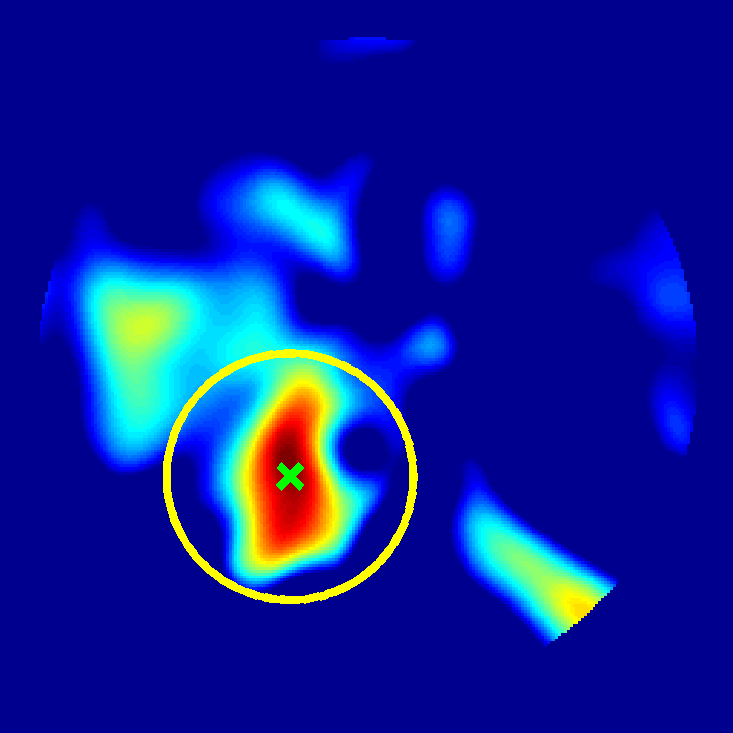}  & 
\includegraphics[width=0.23\textwidth]{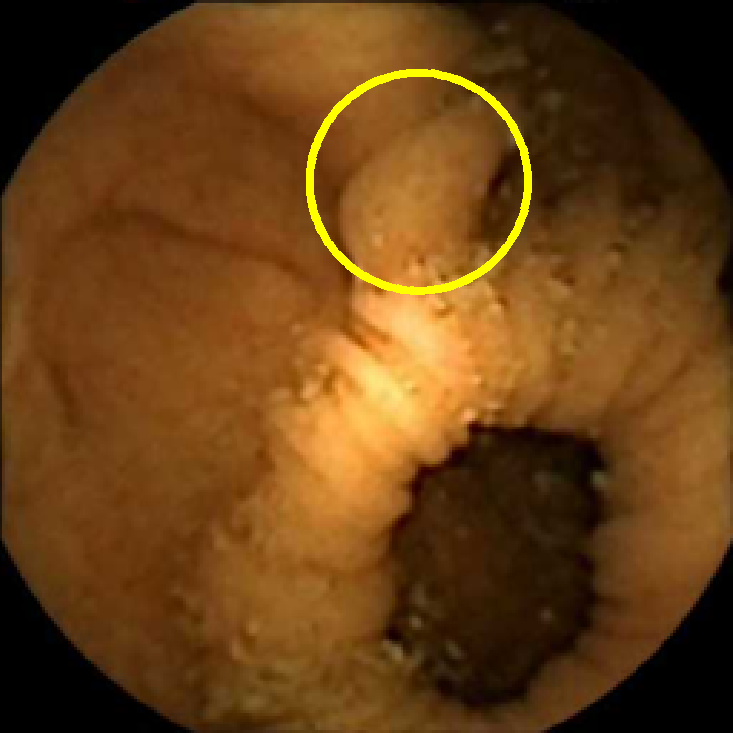} &
\includegraphics[width=0.23\textwidth]{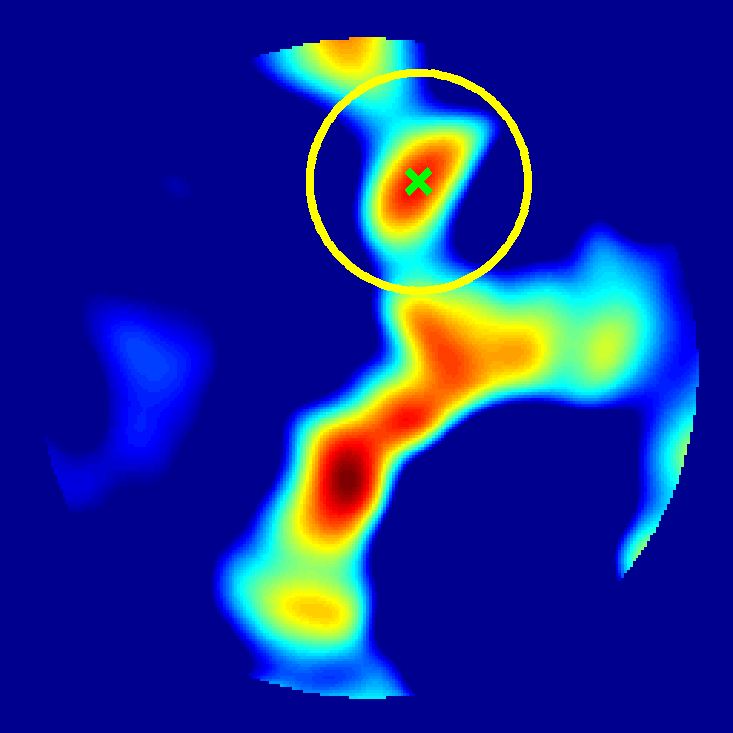} & \\
\includegraphics[width=0.23\textwidth]{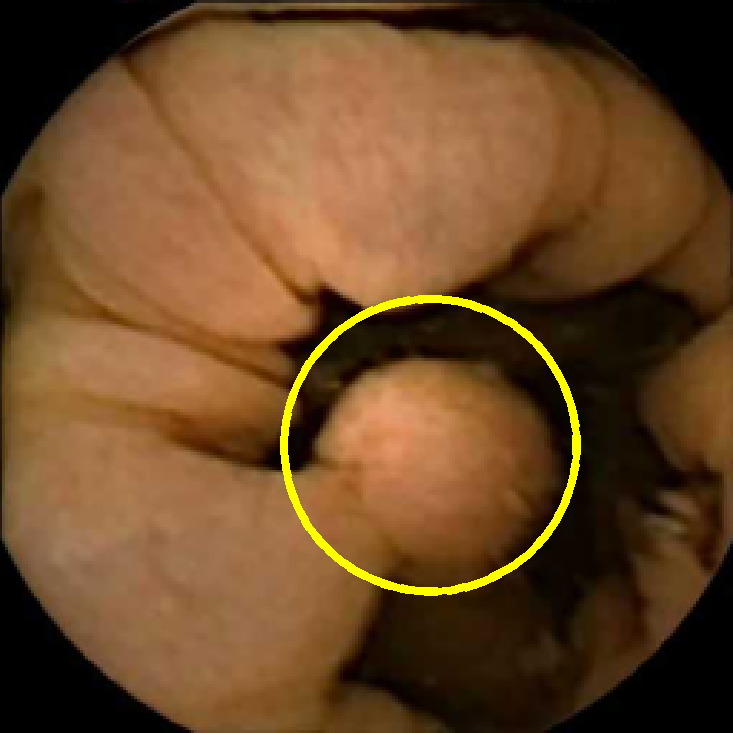} & 
\includegraphics[width=0.23\textwidth]{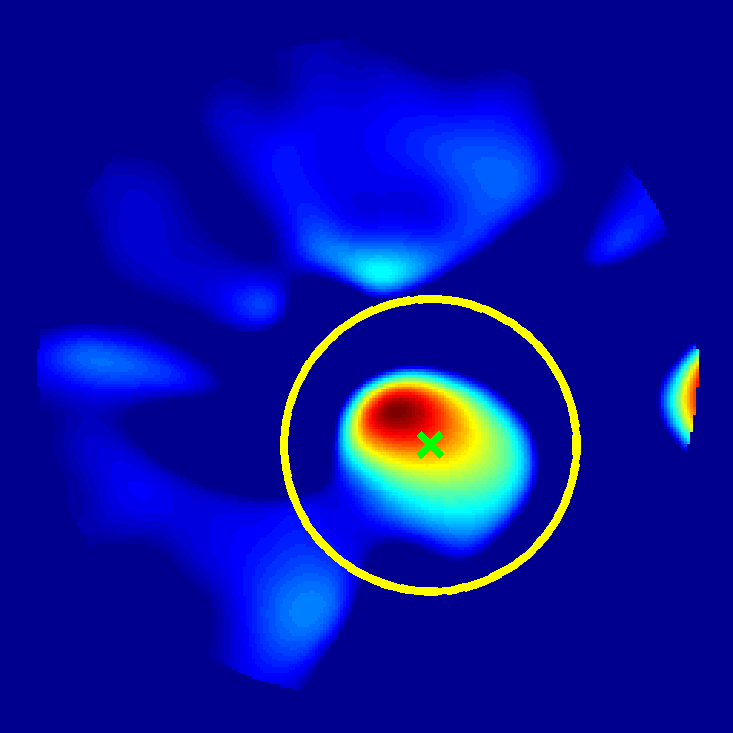} & 
\includegraphics[width=0.23\textwidth]{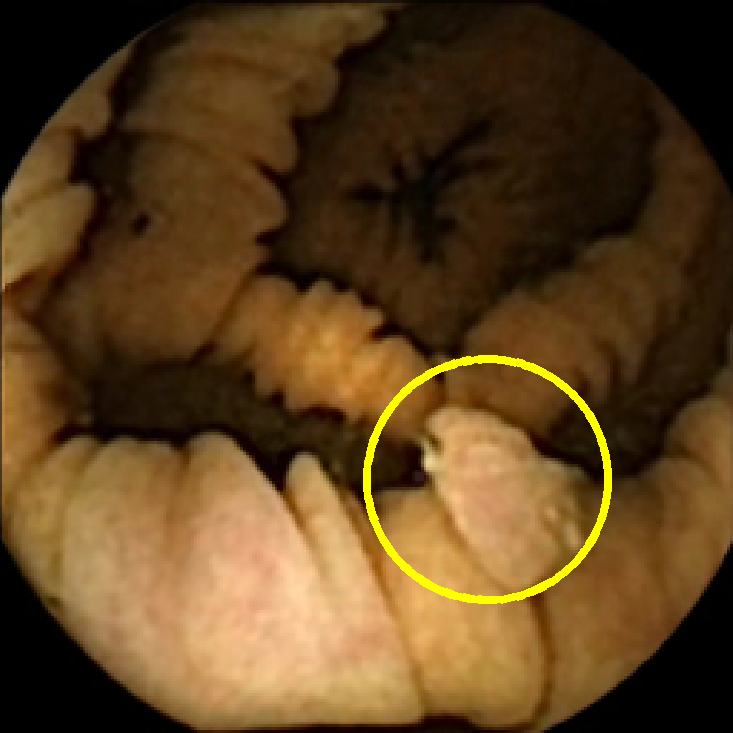} & 
\includegraphics[width=0.23\textwidth]{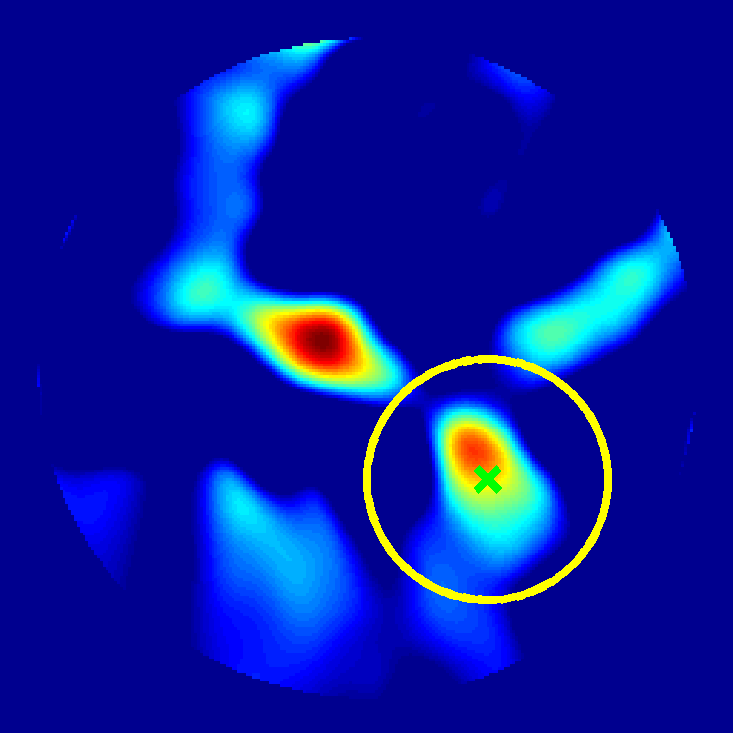} & \\
\includegraphics[width=0.23\textwidth]{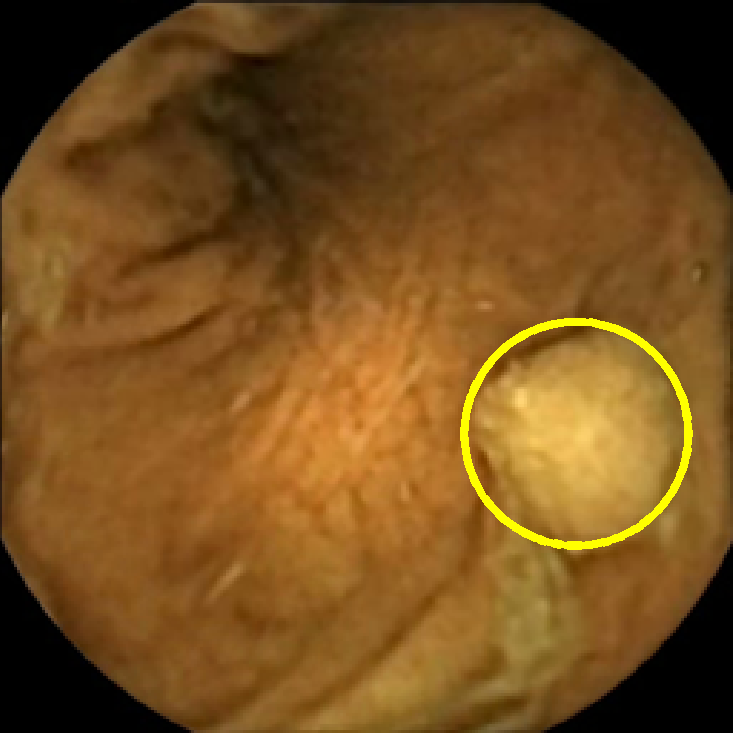} &
\includegraphics[width=0.23\textwidth]{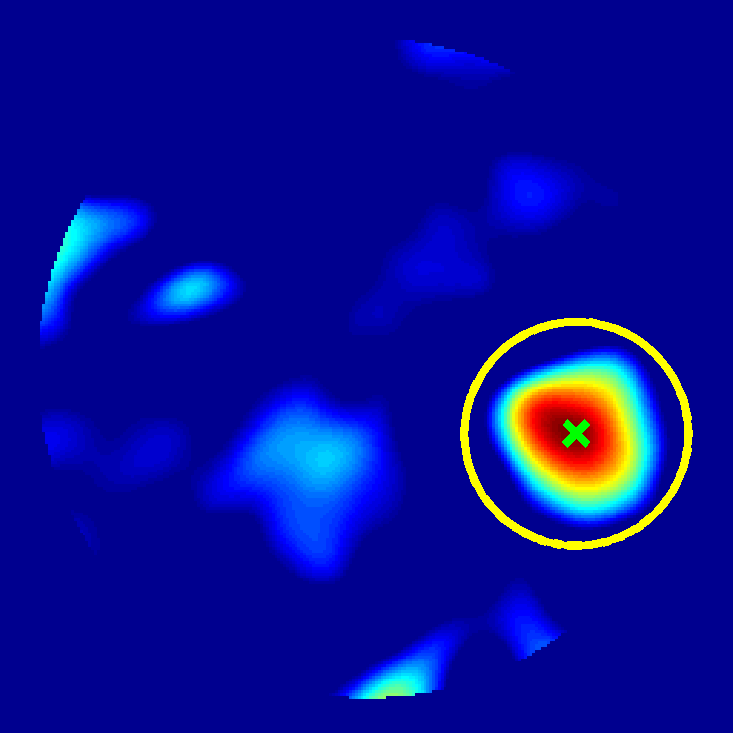} &
\includegraphics[width=0.23\textwidth]{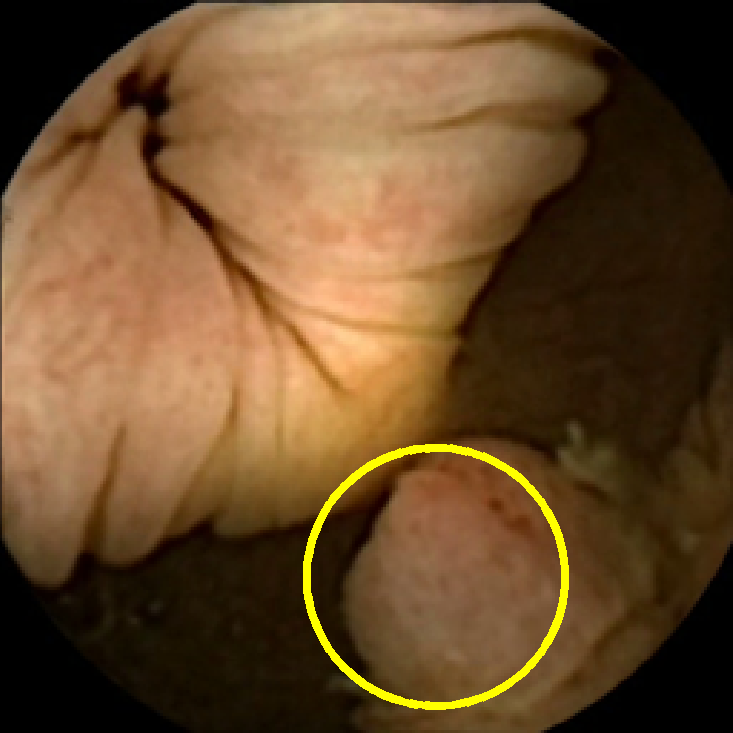} &
\includegraphics[width=0.23\textwidth]{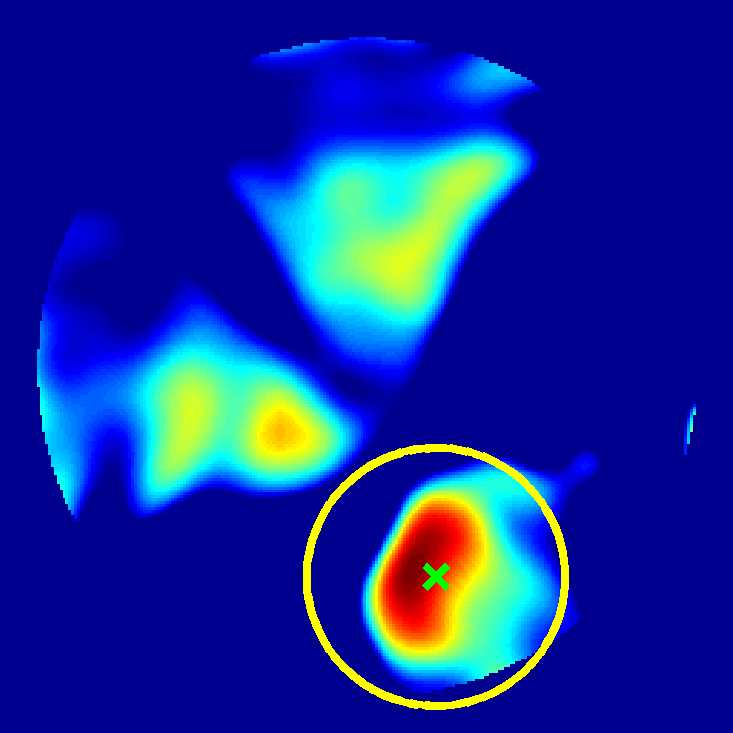} & \\
\centering (a) & \centering (b) & \centering (c) & \centering (d) & 
\end{tabular}
\end{flushleft}
\caption{Optimal fit balls for some of the correctly classified polyps. The circles 
of optimal radii $R_{max}$ (yellow) are superimposed on original polyp frames 
(columns (a) and (c)) and on the corresponding mid-pass filtered images $\bu$ 
(columns (b) and (d)). In columns (b) and (d) the centers of mass 
($\widetilde{c}_x^{(k)}$, $\widetilde{c}_y^{(k)}$) are marked by green $\times$.}
\label{fig:ball}
\end{figure*}

\begin{figure*}
\begin{flushleft}
\begin{tabular}{p{0.22\textwidth}p{0.22\textwidth}p{0.22\textwidth}p{0.22\textwidth}p{0pt}}
\includegraphics[width=0.23\textwidth]{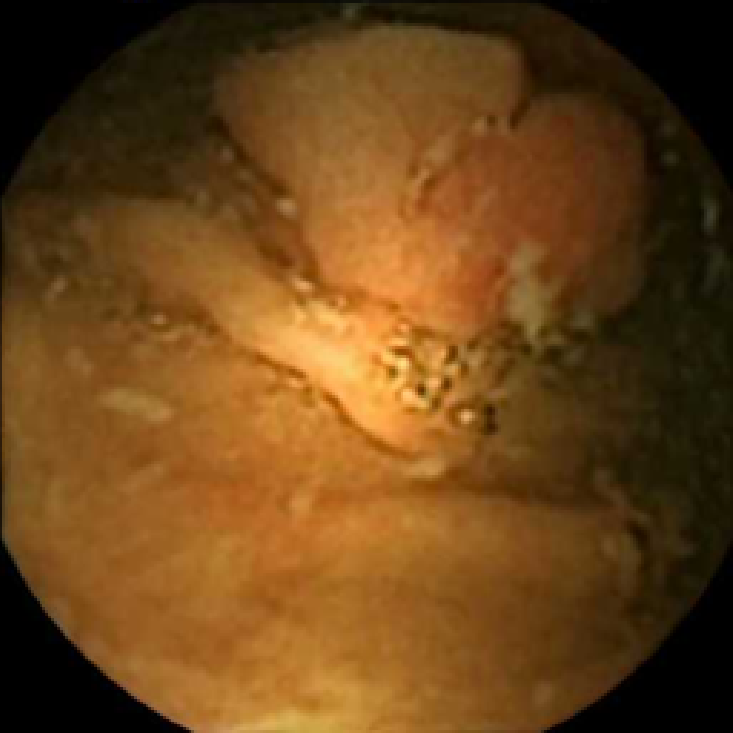}  & 
\includegraphics[width=0.23\textwidth]{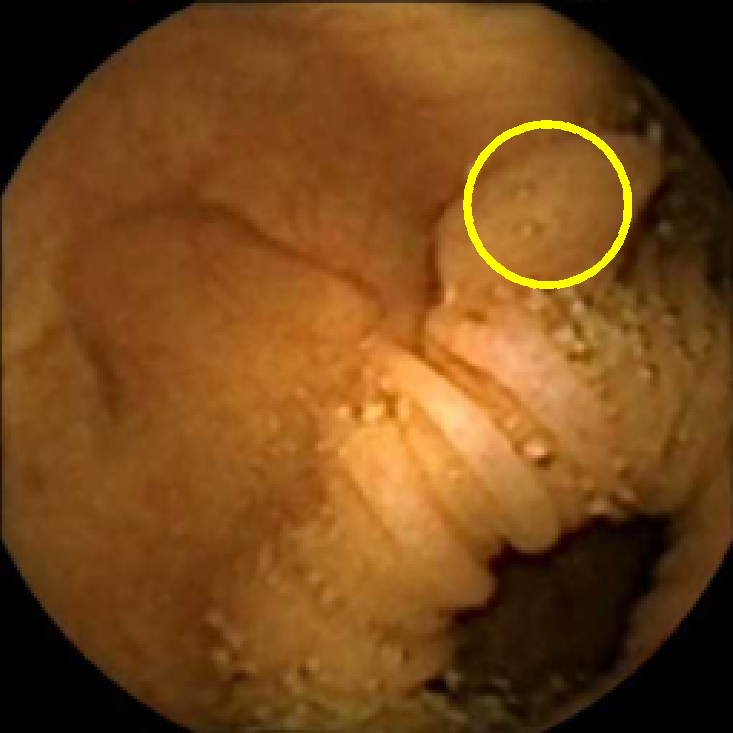}  & 
\includegraphics[width=0.23\textwidth]{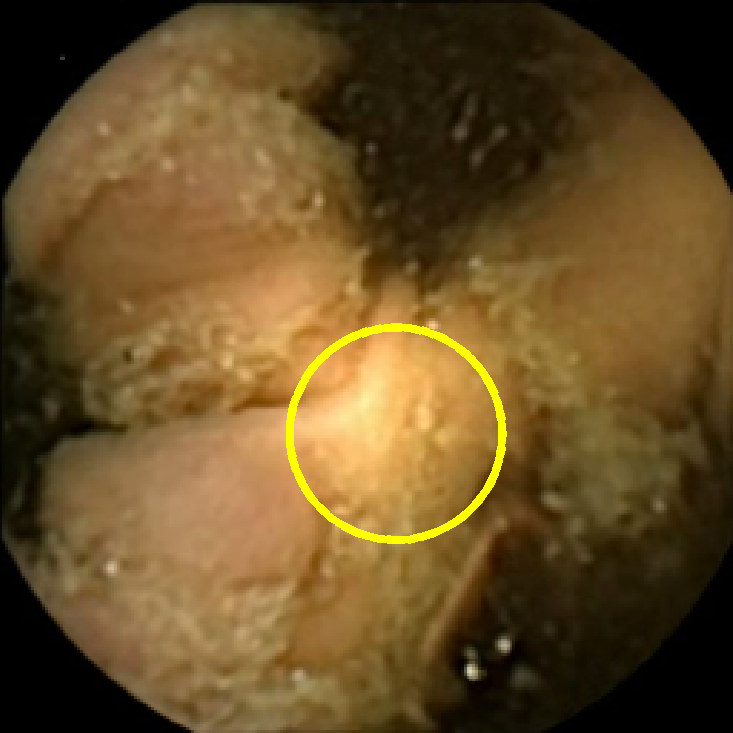}  &
\includegraphics[width=0.23\textwidth]{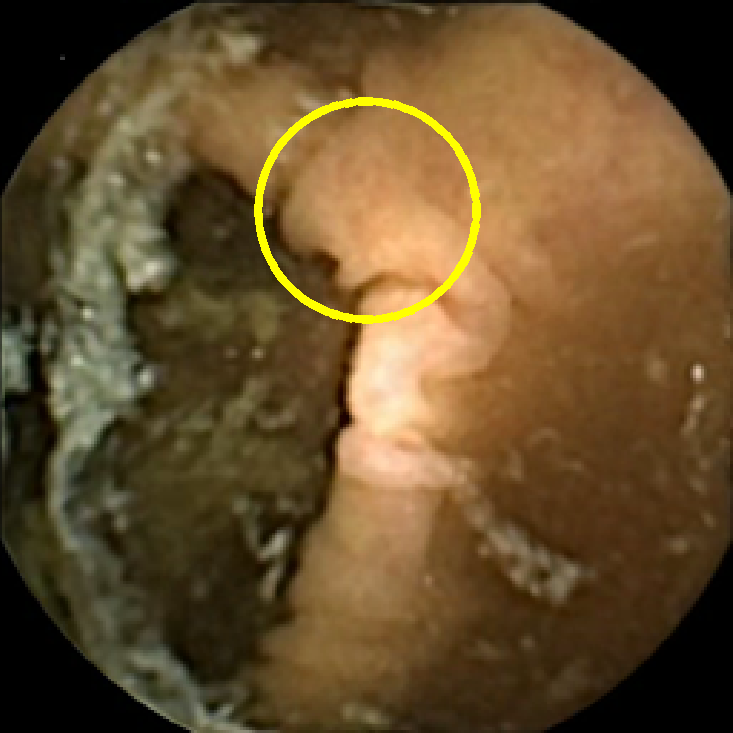}  & \\  
\includegraphics[width=0.23\textwidth]{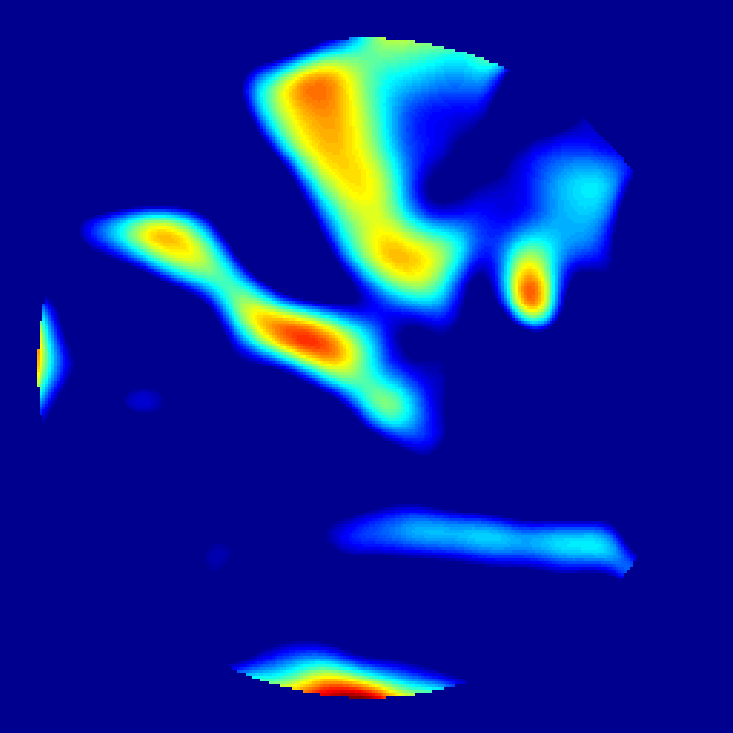}  & 
\includegraphics[width=0.23\textwidth]{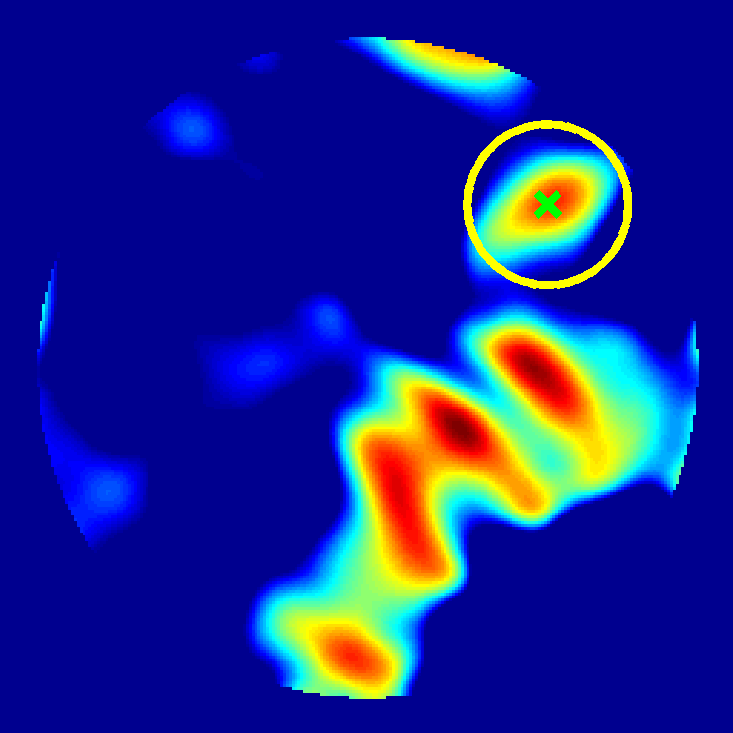}  & 
\includegraphics[width=0.23\textwidth]{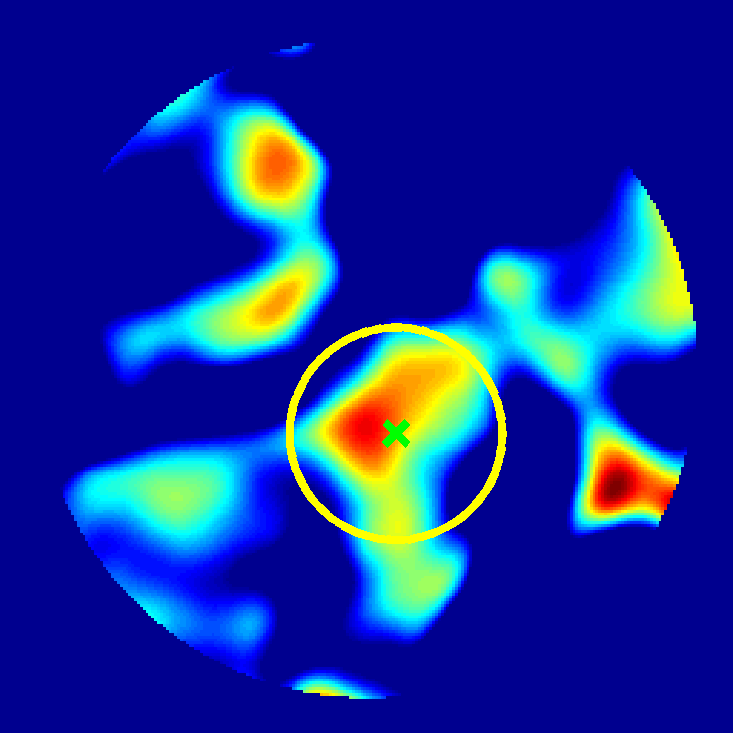}  &
\includegraphics[width=0.23\textwidth]{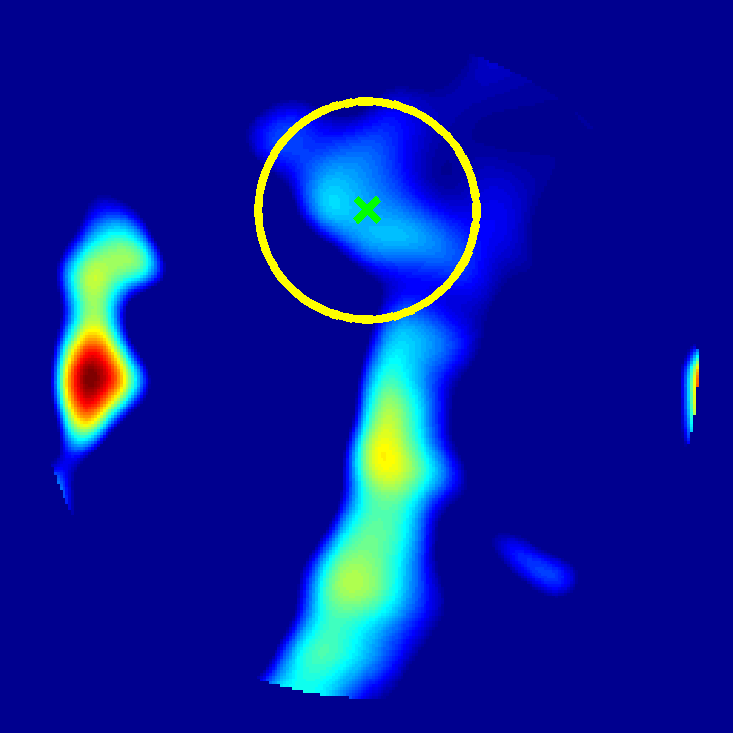}  & \\  
\includegraphics[width=0.23\textwidth]{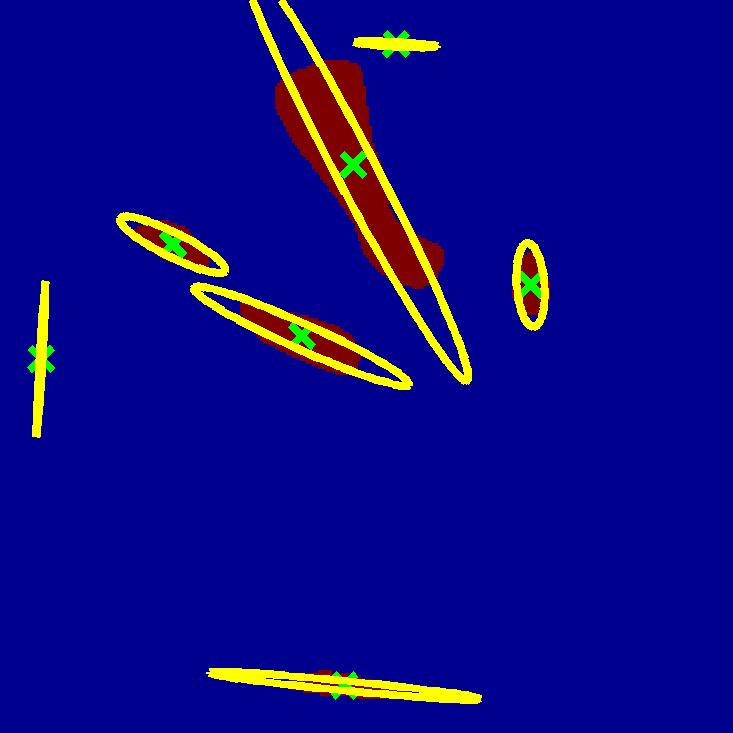}  & 
\includegraphics[width=0.23\textwidth]{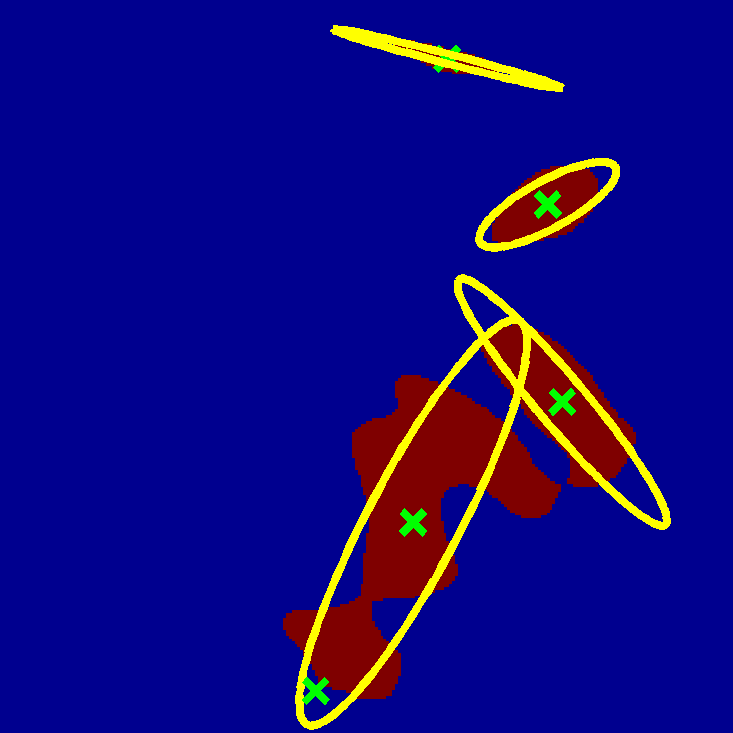}  & 
\includegraphics[width=0.23\textwidth]{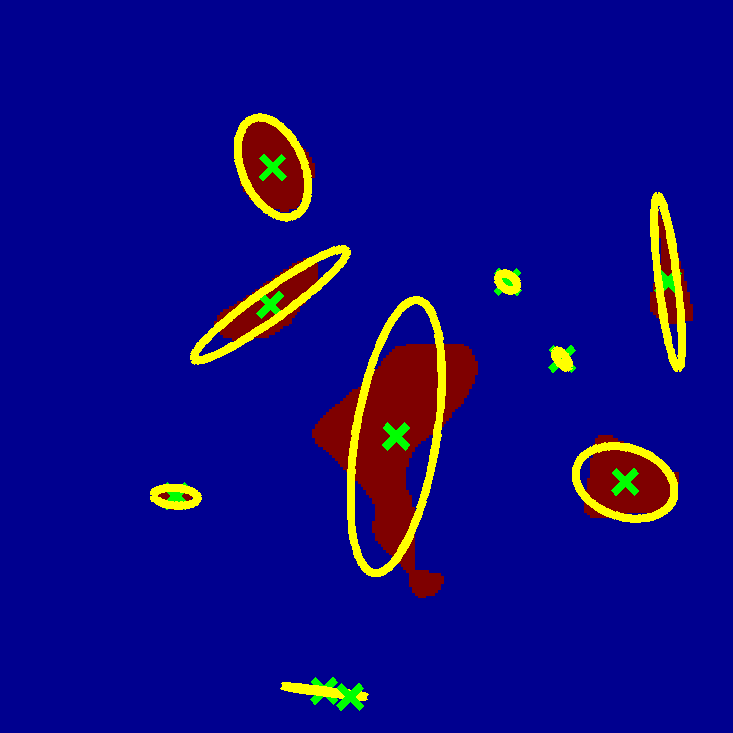}  &
\includegraphics[width=0.23\textwidth]{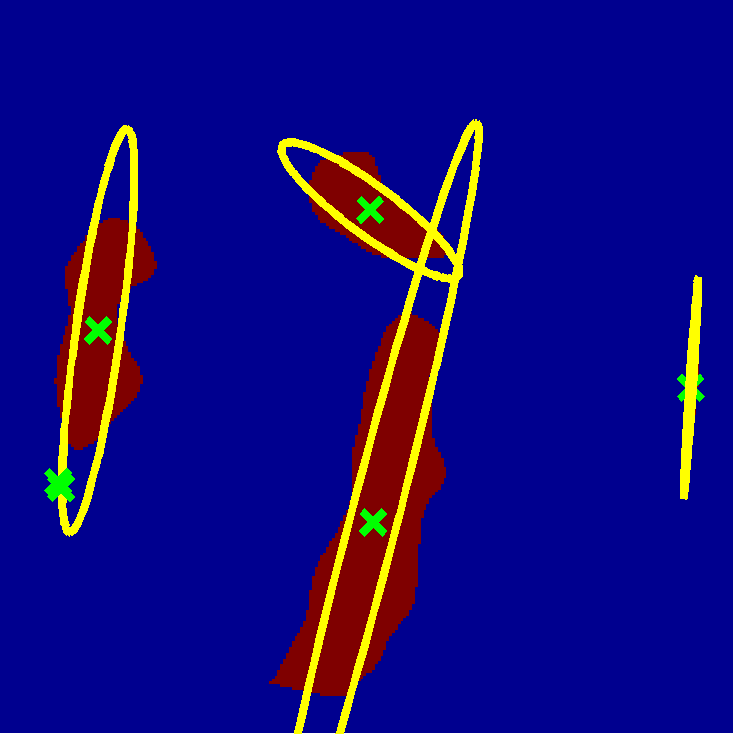}  & \\ 
\centering (a) & \centering (b) & \centering (c) & \centering (d) & 
\end{tabular}
\end{flushleft}
\caption{Incorrectly classified frames. False negatives: columns (a) and (b). 
False positives: columns (c) and (d). 
Top row: original frames with superimposed circles of radii $R_{max}$.
Middle row: mid-pass filtered images $\bu$ with superimposed circles of radii $R_{max}$. 
Bottom row: binary segmentation $\bs = \sum_{k=1}^{N_C} s^{(k)}$ with superimposed 
ellipses of inertia $r^{(k)}(\theta)$.}
\label{fig:fnfp}
\end{figure*}

First, we compute the centers of mass using $\bu$ masked with 
$\bs^{(k)}$ instead of just (\ref{eqn:cmass}), which gives
\begin{equation}
\widetilde{c}_x^{(k)} = \frac{1}{U^{(k)}} \sum_{i,j} x_{ij}^{(k)} u_{ij}, \quad
\widetilde{c}_y^{(k)} = \frac{1}{U^{(k)}} \sum_{i,j} y_{ij}^{(k)} u_{ij},
\end{equation}
for $k \in K_G$, where 
\begin{equation}
U^{(k)} = \sum_{i,j} u_{ij} s_{ij}^{(k)}, \quad k \in K_G,
\end{equation}
and the matrices $x^{(k)}$, $y^{(k)}$ are defined in (\ref{eqn:xy}).

Second, we place a ball with a center at 
$(\widetilde{c}_x^{(k)}, \widetilde{c}_y^{(k)})$ and we search of the radius of
such ball so that it fits best the mid-pass filtered image $\bu$. The radius of
this ball will be the decision parameter in our binary classification. Such 
definition of the decision parameter is motivated by the same considerations
as the criterion (\ref{eqn:geom-criterion}), i.e. we expect the polyps to be
the protrusions that are somewhat rounded. Note that the combined geometric
criterion (\ref{eqn:geom-criterion}) only uses the two-dimensional information
in $\bu$ by only working with the binary segmentation $\bs$. To utilize information
about the height of the protrusions, we need to work with $\bu$ itself, which is why
we fit the ball to $\bu$ instead of $\bs$.

To compute the optimal fit ball radius we define the matrix-valued functions
\begin{equation}
\begin{split}
b_{ij}^{(k)}(R) = & \frac{1}{N_x^2} \left( 
R^2 - (i - \widetilde{c}_y^{(k)})^2 - (j - \widetilde{c}_x^{(k)})^2 \right), \\
& 1 \leq i \leq N_y, \; 1 \leq j \leq N_x,
\end{split}
\end{equation}
and their positive parts
\begin{equation}
\widetilde{\bb}^{(k)}(R) = H(\bb^{(k)}(R)) \cdot \bb^{(k)}(R), \quad k \in K_G,
\label{eqn:btilde}
\end{equation}
where Heaviside step function and the multiplication are performed pixel-wise.

Then for each feature we can define the radius of the ball that provides the best 
fit of $\bu$ as a solution of a one-dimensional optimization problem
\begin{equation}
R_{opt}^{(k)} = \mathop{\mbox{argmin }}\limits_{R} 
\| \bu - \widetilde{\bb}^{(k)}(R) \|_F, \quad k \in K_G,
\label{eqn:ropt}
\end{equation}
where $\|.\|_F$ is the matrix Frobenius norm. Since the objective in the 
optimization problem (\ref{eqn:ropt}) is cheap to evaluate, the problem can be 
easily solved by a simple one-dimensional search over the integer values in 
some interval, which we take here to be $\left[ 1, \lfloor N_x / 3 \rfloor \right]$.

Finally, we can define the decision parameter by taking the maximum of the
optimal fit ball radii over all the features that pass the combined geometric 
criterion (\ref{eqn:geom-criterion}) as
\begin{equation}
R_{max} = \mathop{\mbox{max }}\limits_{k \in K_G} R_{opt}^{(k)}.
\label{eqn:rmax}
\end{equation}
For the frames with $K_G = \varnothing$ we set $R_{max} = 0$. To account for
the pre-selection criterion (\ref{eqn:tcriterion}), we also set
$R_{max} = 0$ for the frames with $T_{max} < T_L$ or $T_{max} > T_U$.

\begin{figure}
\centering
\includegraphics[width=0.45\textwidth]{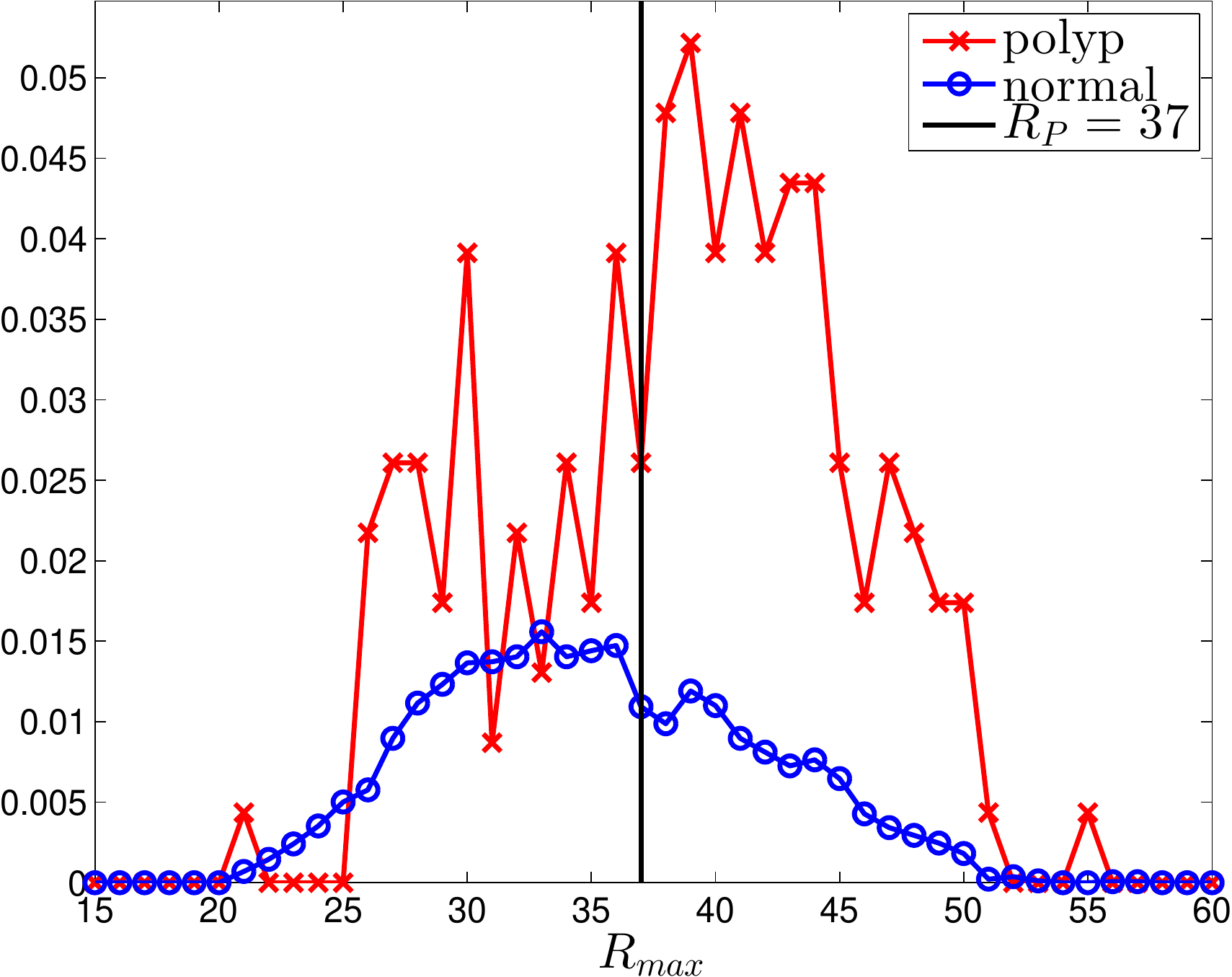}
\caption{Histogram of the distribution of $R_{max}$ for normal and polyp frames 
after passing the pre-selection (\ref{eqn:tcriterion}) and satisfying the combined 
geometric criterion (\ref{eqn:geom-criterion}).}
\label{fig:hist-radius}
\end{figure}

Since we expect the polyps to correspond to more pronounced round protrusions,
we define the binary classifier as
\begin{equation}
BC(\bff) = \left\{
\begin{tabular}{ll}
``polyp'', & if $R_{max} \geq R_P$ \\
``normal'', & if $R_{max} < R_P$
\end{tabular}
\right..
\label{eqn:binclass}
\end{equation}
Obviously, the performace of the classifier depends dramatically on the choice
of the \emph{discrimination threshold} value $R_P$. This choice is discussed 
in detail in section \ref{subsec:roc}, where we use statistical analysis to
determine the value of $R_P$ that provides the desired performance.

In Figure \ref{fig:ball} we show the circles of radius $R_{max}$ corresponding
to the features that were correctly classified as polyps by (\ref{eqn:binclass}).
We observe that the classifier was able to identify the polyps of a variety 
of shapes even in the presence of small amounts of trash liquid (first row) or
when the polyps are located next to mucosal folds (rows two to four in column (c)).

The examples of incorrect classification of frames are presented in 
Figure \ref{fig:fnfp}. The first two examples show false negatives, each highlighting
a possible source of classification error. The example in column (a) shows the
case where the feature corresponding to the polyp was too stretched out and thus
was rejected by the eccentricity criterion (\ref{eqn:ecc-criterion}).
In contrast, the feature corresponding to the polyp in column (b) has passed the
combined geometric criterion (\ref{eqn:geom-criterion}), but the radius $R_{max}$
was below the threshold $R_P=37$ of the binary classifier. Examples in columns
(c) and (d) show the two sources of possible false positives. The false positive
detection in column (c) is due to insufficient illumination correction. The bright 
spot is not fully corrected at the pre-processing stage and subsequently generates 
a polyp-like feature in the mid-pass filtered frame that happens to pass through 
all the criteria. Finally, in column (d) a mucosal fold is classified as polyp. 
Note that such cases are the most difficult to deal with, as the mucosal
folds can often be hard to distinguish from polyps even for a human operator.

The overall good performance of the classifier can be explained by statistical 
considerations. In Figure \ref{fig:hist-radius} we show the histogram of the 
distribution of $R_{max}$ for normal and polyp frames in our test data set
(see section \ref{subsec:dataset} for the detailed description). We observe that
the peak of the distribution of $R_{max}$ for polyp frames is shifted to the 
right compared to the peak of the distribution for ``normal'', non-polyp frames.
Note that overall the distribution for normal frames is well below the distribution
for the polyp frames. This is due to setting $R_{max}$ to zero for the frames 
that do not pass the pre-selection or that do not satisfy the combined geometric 
criterion.

\subsection{Summary of the algorithm}
\label{subsec:sumalg}

\begin{figure*}
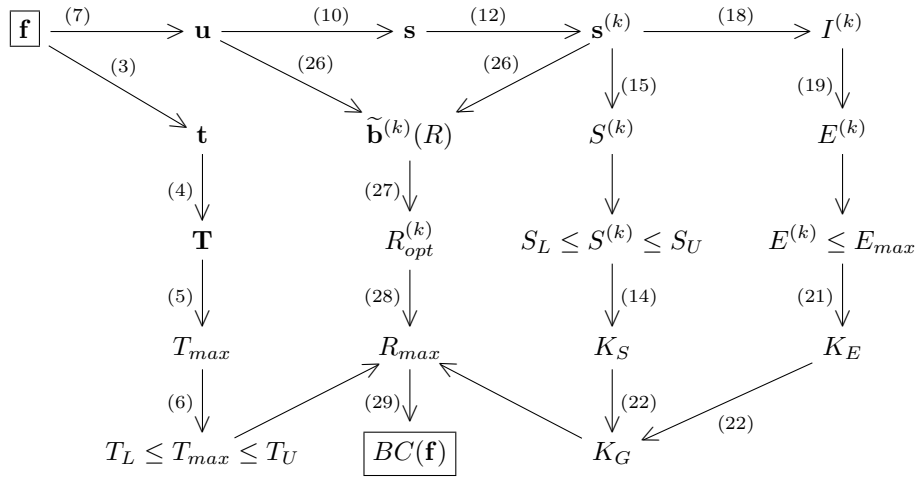

\begin{diagram}
\boxed{\bff} & \rTo^{(\ref{eqn:midpass})} & \bu & \rTo^{(\ref{eqn:segment})} & \bs & 
\rTo^{(\ref{eqn:conncomp})} & \bs^{(k)} & \rTo^{(\ref{eqn:tensor})} & I^{(k)} \\
 & \rdTo^{(\ref{eqn:tc})} & & \rdTo^{(\ref{eqn:btilde})} & & \ldTo^{(\ref{eqn:btilde})} & 
\dTo_{(\ref{eqn:size})} & & \dTo^{(\ref{eqn:ecc})} \\
 & & \bt & & \widetilde{\bb}^{(k)}(R) & & S^{(k)} & & E^{(k)} \\
 & & \dTo^{(\ref{eqn:tconv})} & & \dTo^{(\ref{eqn:ropt})} & & \dTo & & \dTo \\
 & & \bT & & R^{(k)}_{opt} & & S_L \leq S^{(k)} \leq S_U & & E^{(k)} \leq E_{max} \\
 & & \dTo^{(\ref{eqn:tmax})} & & \dTo^{(\ref{eqn:rmax})} & & \dTo_{(\ref{eqn:size-criterion})} & & 
\dTo^{(\ref{eqn:ecc-criterion})} \\
 & & T_{max} & & R_{max} & & K_S & & K_E \\
 & & \dTo^{(\ref{eqn:tcriterion})} & \ruTo & \dTo^{(\ref{eqn:binclass})} & \luTo & 
\dTo_{(\ref{eqn:geom-criterion})} & \ldTo_{(\ref{eqn:geom-criterion})} & \\
 & & T_L \leq T_{max} \leq T_U & & \boxed{BC(\bff)} & & K_G & &
\end{diagram}
\caption{The flow of the data in Algorithm \ref{alg:class}.}
\label{fig:dataflow}
\end{figure*}

After establishing the main steps of the algorithm, we are ready to summarize
the processing of a single frame in a video sequence. The processing algorithm 
accepts the frame as an input and gives the classification ``polyp'' or ``normal'' 
as an output. The flow of data in the algorithm is illustrated in
Figure \ref{fig:dataflow}.
\begin{alg}[Binary classification with pre-selection]~~
\label{alg:class}
\begin{enumerate}
\item \textup{\textbf{Pre-processing}}. Convert the frame to grayscale, perform the
intensity normalization, apply the linear extension outside the circular mask
of radius $R_{mask}$ to obtain the pre-processed frame $\bff$.
\item \textup{\textbf{Texture analysis}}. \label{step:text} Compute the texture $\bt$ from (\ref{eqn:tc}) and its 
non-linear convolution-type transform $\bT$ using (\ref{eqn:tconv}). Find the 
maximum $T_{max}$ of $\bT$ and apply the pre-selection criterion 
(\ref{eqn:tcriterion}). If the frame fails the criterion, set $R_{max}=0$ and 
go to step \ref{step:class}, otherwise continue.
\item \textup{\textbf{Mid-pass filtering and segmentation}}. 
Apply the mid-pass filter (\ref{eqn:midpass}) to obtain $\bu$ and 
perform the binary segmentation (\ref{eqn:segment}) of $\bu$ to obtain $\bs$.
Decompose the binary image $\bs$ into connected components $\bs^{(k)}$, 
$k=1,\ldots,N_C$ that correspond to $N_C$ ``features'' present in the frame.
\item \textup{\textbf{Geometric analysis}}. \label{step:geom} 
For each of $N_C$ features compute the tensor of 
inertia $I^{(k)}$ via (\ref{eqn:tensor}) and the eccentricity $E^{(k)}$ of 
the corresponding ellipse of intertia (\ref{eqn:ecc}). Apply the eccentricity 
criterion (\ref{eqn:ecc-criterion}) and the feature size criterion 
(\ref{eqn:size-criterion}) to obtain the features $K_G$ that satisfy both
criteria. If $K_G = \varnothing$, set $R_{max}=0$ and go to step \ref{step:class},
otherwise continue.
\item \textup{\textbf{Ball fitting}}.
For each of the features passing the combined geometric criterion 
(\ref{eqn:geom-criterion}) compute the radius $R^{(k)}_{opt}$ of the best fit
ball. Take the maximum $R_{max}$ of these radii over $k \in K_G$ (\ref{eqn:rmax}). 
\item \textup{\textbf{Final binary classification}}.
\label{step:class} Apply the binary classifier (\ref{eqn:binclass}) to 
$R_{max}$ to classify the frame as either ``normal'' or ``polyp''.
\end{enumerate}
\end{alg}

Note that the algorithm is quite inexpensive computationally. The computational
cost of processing a single frame is of the order of $O(N_x N_y)$ operations. 
None of the steps requires a solution of expensive PDE or optimization problems, 
that are often used in image processing. The only minimization sub-problem is a 
simple one-dimensional search (\ref{eqn:ropt}), which can be done very efficiently. 
Even with a very crude Matlab implementation, the algorithm takes less than 
one second per frame on a regular desktop. Obviously, one would expect a 
proper C/C++ implementation to be a lot more efficient. This gives an 
advantage to our algorithm in the common situations where the captured video
sequence contains thousands of frames, which makes the processing time an
important issue.

\section{Testing methodology and data set}
\label{sec:method}

In this section we discuss the methodology of a statistical test of performance 
of Algorithm \ref{alg:class} and the testing data set. The algorithm's 
implementation and all the computations were performed with Matlab and the Image 
Processing Toolbox.

\subsection{Data set}
\label{subsec:dataset}

A key to developing an efficient and robust algorithm for polyp detection is 
being able to test it on a sufficiently rich data set. Using a small number of
sample frames can easily lead to overtuning that may create an illusion of 
good performance, but in realistic conditions such an algorithm can easily fail.

In this work we were able to use a data set courtesy of the Hospital of 
the University of Coimbra. The data set contains $N = 18968$ frames, out of which 
$N_N = 18738$ are normal frames and $N_P = 230$ frames contain polyps. The 
frames are taken from the full exam videos of five adult patients. The videos 
were captured with PillCam COLON 2 capsule (manufactured by Given Imaging, 
Yoqneam, Israel, \url{http://www.givenimaging.com}) in the native resolution of 
$512 \times 512$ pixels and were downsampled to $N_x = N_y = 256$ before 
processing. The downsampling was performed to reduce the processing time for
each frame. Since the mid-pass filter applied to the frames is smoothing in 
nature, the downsampling has minimal effect on the performance of the algorithm.

The ``normal'' part of the data set contains frames with mucosal folds, 
diverticula, bubbles and trash liquids, which allows us to test our algorithm 
in realistic conditions. The sheer number of non-polyp frames in the data set 
ensures that our algorithm not only has a high sensitivity (high true positive 
rate), but also high specificity (low false positive rate).

The part of the data set containing the frames with polyps is organized into
sequences corresponding to each polyp. A total of $16$ polyps are present 
in 230 ``polyp'' frames. The lengths of these sequences are given in Table
\ref{tab:seq}. Grouping the frames into sequences allows us to study the
performance of the algorithm not only on \emph{per frame} basis, but also on
\emph{per polyp} basis. The second row of Table \ref{tab:seq} presents the
results of such study, which is described in detail in the next section.

\subsection{Receiver operating characteristic curve}
\label{subsec:roc}

To properly test an algorithm of polyp detection it is not enough to just try 
it on a limited number of specially chosen frames. The test should be 
statistical in nature, which is the reason for using a data set with over 
$18,900$ frames. The measures of performance should also capture the 
statistical nature of the testing, which leads us to the consideration of 
\emph{receiver operator characteristic} also referred to as the \emph{ROC curve}.

ROC curves are a standard tool for evaluating the performance of binary 
classifiers. They quantify the change in performance of a classifier (in our 
case $BC(\bff)$ defined in (\ref{eqn:binclass})) as the discrimination threshold 
value (in our case $R_P$) is varied. To define the ROC curve we need to 
introduce the following quantities. Suppose that we know a true classification 
$TC(\bff)$ of every frame in the data set as either ``polyp'' or ``normal''. Then 
we can define the true positive rate $TPR$ and the false positive rate $FPR$ as
\begin{eqnarray}
TPR & = & \frac{1}{N_P}
\left(\begin{tabular}{l}
number of frames $\bff$ s.t.\\
$TC(\bff)=BC(\bff)=$``polyp''
\end{tabular}\right),
\label{eqn:tpr-frame} \\
FPR & = & \frac{1}{N_N}
\left(\begin{tabular}{l}
number of frames $\bff$ s.t. \\
$TC(\bff)=$``normal'', \\
but $BC(\bff)=$``polyp''
\end{tabular}\right).
\end{eqnarray}
The true positive rate is also known as \emph{sensitivity}
\begin{equation}
SENS = TPR \cdot 100\%,
\end{equation}
which measures the likelihood of the classifier correctly labeling the frame 
containing a polyp as ``polyp''. The false positive rate is used to define the 
\emph{specificity}
\begin{equation}
SPEC = (1-FPR) \cdot 100\%,
\end{equation}
which measures how likely the classifier is to correctly label a non-polyp 
frame as ``normal''.

Obviously, we would like both the sensitivity and the specificity to be as 
high as possible. However, there is always a trade off between the two. To 
visually represent this trade off we use the ROC curve. It is a parametric 
curve in the space $(FPR, TPR)$, where the changing parameter is the 
discrimination threshold $R_P$. A ROC curve connects the points $(0,0)$ and 
$(1,1)$. If the classifier makes a decision randomly with equal probability, 
then the ROC curve will simply be a diagonal $TPR = FPR$. For any classifier 
that behaves better we expect to have a concave ROC curve that deviates far 
from the diagonal. 

We use the ROC curves to asses the performance of Algorithm 
\ref{alg:class} in the following way. We choose a \emph{training} subset
of the whole data set. Here we take the subset corresponding to Patient $4$
with over $8,500$ frames (see Table \ref{tab:falsepos} for individual 
patients' frame counts). We set a desired high level of specificity, e.g. 
$SPEC = 90\%$. Then we compute the values of the decision parameter $R_{max}$ 
for all the frames in the training subset. Among all values of the threshold 
$R_P$ we choose a minimal value that provides at least that much specificity. 
For this value of $R_P$ we compute the specificity and sensitivity of the 
binary classifier for the whole data set. To put these values of specificity
and sensitivity into context we plot them on the ROC curve computed for the 
whole data set.

In the clinical setting the main purpose of the automated processing of 
the capsule endoscope videos is not to detect the individual polyp 
frames, but to find the actual polyps. Typically one polyp will be 
visible not just in a single frame, but in a sequence of frames, even 
for the cameras with low frame rates. Since the frames classified by the 
algorithm as ``polyp'' are going to be inspected manually afterward, 
it is enough for a single frame in the sequence to be classified as 
``polyp'' for the actual polyp to be detected.

To measure the sensitivity of the algorithm to actual polyps instead of 
the single polyp frames, we can use a modified definition of $TPR$. For 
each of $N_S$ sequences corresponding to a single polyp we define the 
detection flags
\begin{equation}
D^{(p)} = \left\{ 
\begin{tabular}{ll}
1, & \begin{tabular}{l} if for at least one $\bff$ in the $p^{th}$ sequence\\ 
$BC(\bff)=$``polyp'' \end{tabular} \\
0, & \begin{tabular}{l} if for all $\bff$ in the $p^{th}$ sequence\\ 
$BC(\bff)=$``normal''\end{tabular}
\end{tabular}
\right.
\end{equation}
Then we can define a \emph{per polyp} $TPR$ as
\begin{equation}
TPR = \frac{1}{N_S} \sum\limits_{p=1}^{N_S}D^{(p)}.
\label{eqn:tpr-polyp}
\end{equation}
We also call the corresponding sensitivity and the ROC curve as defined 
on a \emph{per polyp} basis. This is to distinguish from those based on 
(\ref{eqn:tpr-frame}), that we may refer to as being defined on a
\emph{per frame} basis. 

\subsection{The choice of parameters and robustness}
\label{subsec:robust}

\begin{table*}
\caption{Nomenclature and values of the parameters used in the numerical experiments.}
\centering
\begin{tabular}{c|c|p{0.56\textwidth}|c}
Parameter & Value & Description & Section \\ \hline
$N_x$ & 256 & \raggedright Width of the frame in pixels &  \ref{subsec:dataset} \\ \hline
$N_y$ & 256 & \raggedright Height of the frame in pixels & \ref{subsec:dataset} \\ \hline
$R_{mask}$ & $0.45 \, N_x$ & 
\raggedright Radius of a circular mask in pixels & \ref{subsec:pre} \\ \hline
$n_{iter}$ & 5 & 
\raggedright Number of iterations for the texture+cartoon decomposition (\ref{eqn:tc}) & 
\ref{subsec:texture} \\ \hline
$\sigma_t$ & 5 & 
\raggedright Gaussian standard deviation for the texture+cartoon decomposition (\ref{eqn:tc}) & 
\ref{subsec:texture} \\ \hline
$\sigma$ & $\lceil N_x/25 \rceil$ & 
\raggedright Gaussian standard deviation for the texture convolution-type transform (\ref{eqn:tconv}) & 
\ref{subsec:texture} \\ \hline
$p$ & 0.8 & 
\raggedright Power for the texture convolution-type transform (\ref{eqn:tconv}) & 
\ref{subsec:texture} \\ \hline
$T_L$ & 3 & 
\raggedright Pre-selection criterion lower threshold (\ref{eqn:tcriterion}) & 
\ref{subsec:texture} \\ \hline
$T_U$ & 8 & 
\raggedright Pre-selection criterion upper threshold (\ref{eqn:tcriterion}) & 
\ref{subsec:texture} \\ \hline
$\sigma_1$ & 7 & 
\raggedright Lower standard deviation in the mid-pass filter (\ref{eqn:midpass}) & 
\ref{subsec:midpass} \\ \hline
$\sigma_2$ & 30 & 
\raggedright Upper standard deviation in the mid-pass filter (\ref{eqn:midpass}) & 
\ref{subsec:midpass} \\ \hline
$M_L$ & 0.11 & 
\raggedright Lower bound for the segmentation threshold (\ref{eqn:theta}) & 
\ref{subsec:midpass} \\ \hline
$M_U$ & 0.16 & 
\raggedright Upper bound for the segmentation threshold (\ref{eqn:theta}) & 
\ref{subsec:midpass} \\ \hline
$S_L$ & $\lceil \left( N_x/15 \right)^2 \rceil$ & 
\raggedright Lower threshold for the feature size criterion (\ref{eqn:size-criterion}) & 
\ref{subsec:geomproc} \\ \hline
$S_U$ & $\lceil \left( N_x/4.5 \right)^2 \rceil$ & 
\raggedright Upper threshold for the feature size criterion (\ref{eqn:size-criterion}) &
\ref{subsec:geomproc} \\ \hline
$E_{max}$ & 6.5 & 
\raggedright Maximum eccentricity of the ellipse of inertia for the criterion (\ref{eqn:ecc-criterion}) & 
\ref{subsec:geomproc} \\ \hline
$R_P$ & 37 & 
\raggedright Discrimination threshold in the binary classifier (\ref{eqn:binclass}) &
\ref{subsec:ball} 
\end{tabular}
\label{tab:params}
\end{table*}

Algorithm \ref{alg:class} relies on certain numerical parameters. A complete list 
of these parameters and their values used in the numerical experiments is given in 
Table \ref{tab:params}. Most of the parameters are of geometric nature, i.e.
they relate to the size or shape of certain features in the frame. Their values
were chosen manually based on the expectations of the size and shape of the polyps
that the method is likely to encounter. 

While choosing the parameters we made the best effort to avoid any fine-tuning. 
The values were chosen from the common sense considerations, not from the 
considerations of improving the performance for a particular data set that we used.
Since we do not have an automated procedure for choosing all the parameters in 
a way optimal in some sense, we cannot assess the best possible performance of the
algorithm. However, we can perform a robustness study. If the algorithm is
robust enough with respect to the changes in the parameters, we can expect that
the performance demonstrated with a current choice of parameters is not far 
from the optimal performance over all possible parameter values.

To estimate the robustness of Algorithm \ref{alg:class} we compute the 
\emph{sensitivities}, denoted by $\delta(\cdot)$, of the statistical quantities
$SPEC$ and $SENS$ both on per frame and per polyp basis. The sensitivities
$\delta(\cdot)$ are not to be confused with the sensitivity $SENS$ of the binary
classification. The sensitivities are computed with respect to one parameter
at a time. Suppose we want to compute the sensitivity of $SPEC$ with respect
to some parameter $X$. First, we choose the base value $X_{base}$ and then a
perturbed value $X_{pert}$. Second, we compute the values of $SPEC$ for both
values of $X$ while keeping all other parameters fixed. We denote these values
by $SPEC_{base}$ and $SPEC_{pert}$ respectively. Finally, the sensitivity 
of $SPEC$ with respect to $X$ is defined by
\begin{equation}
\delta(SPEC) = \frac{| SPEC_{pert} - SPEC_{base} |}{SPEC_{base}} \cdot 100 \%.
\label{eqn:deltaSPEC}
\end{equation}
Note that we work with the relative sensitivities here. 

For the purpose of the robustness study we took the base values of the parameters
as in Table \ref{tab:params}. The perturbed values were taken at $10\%$ larger 
than the base values, i.e. $X_{pert} = 1.1 \cdot X_{base}$. Where the parameters
assume integer values (e.g. $\sigma_1$, $\sigma_2$, $S_L$, $S_U$), the resulting 
number was rounded up. Such a substantial perturbation of the base value ensures 
that the robustness study takes into account the full non-linearity of the algorithm. 

Since the sensitivity calculation with respect to each parameter requires a
full statistical calculation, we performed it on a reduced data set for the
purpose of reducing the required computational effort. While we kept all
polyp frames, we reduced the number of normal frames to $4000$. Also, we restricted
the number of parameters with respect to which the sensitivities were computed. 
Since the bulk of the algorithm is the geometric part (steps 3-6) and not the 
pre-selection (step 2), we restrict the robustness study to the parameters
that enter the geometric part.

\section{Testing results}
\label{sec:results}

In this section we provide the results of a statistical test of Algorithm 
\ref{alg:class} according to the methodology outlined in section \ref{sec:method}. 

\subsection{Sensitivity and specificity}
\label{subsec:resroc}

\begin{figure}
\centering
\begin{tabular}{p{0.45\textwidth}p{0pt}}
\includegraphics[width=0.4\textwidth]{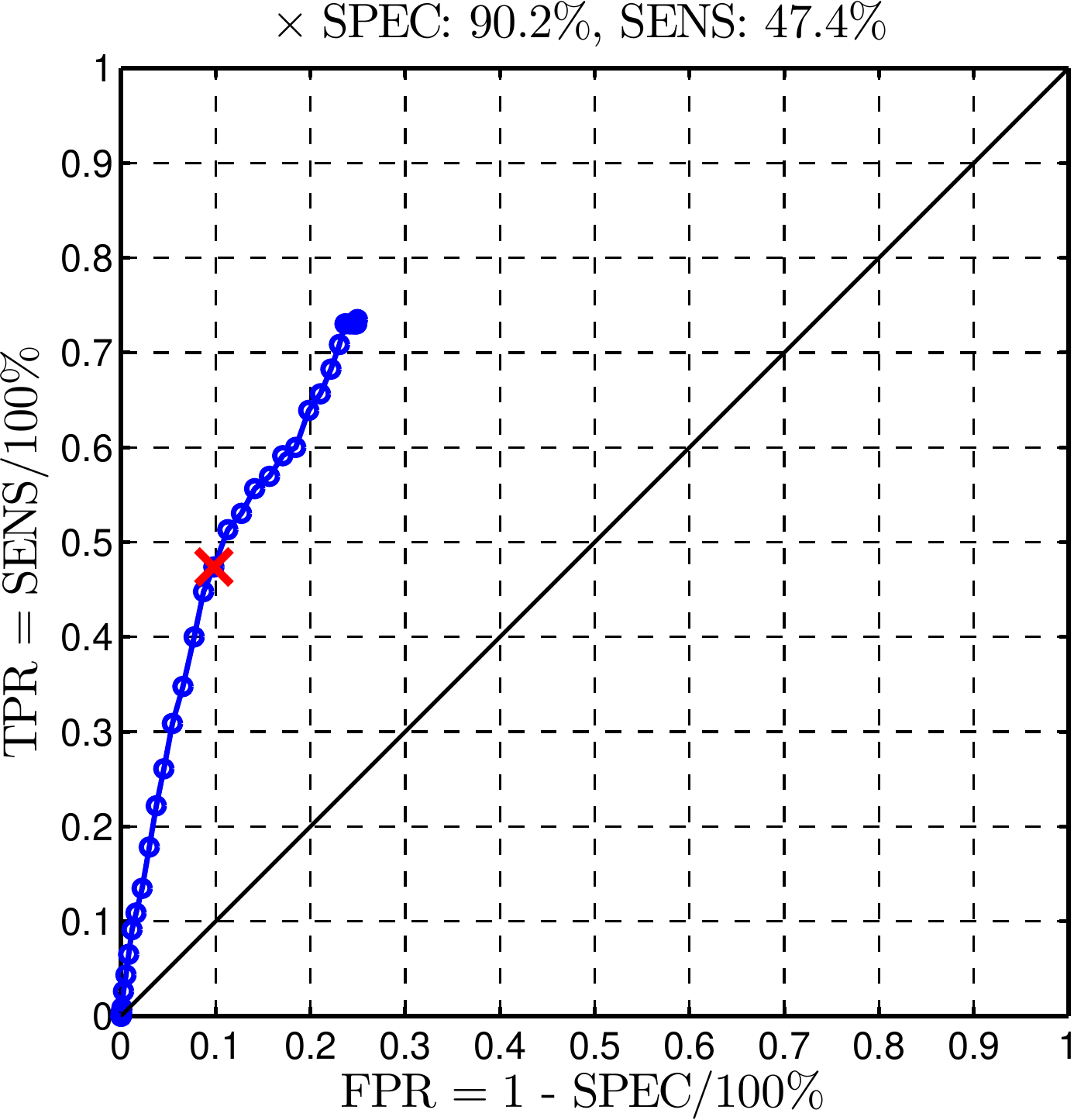} & \\
\centering (a) & \\
\includegraphics[width=0.4\textwidth]{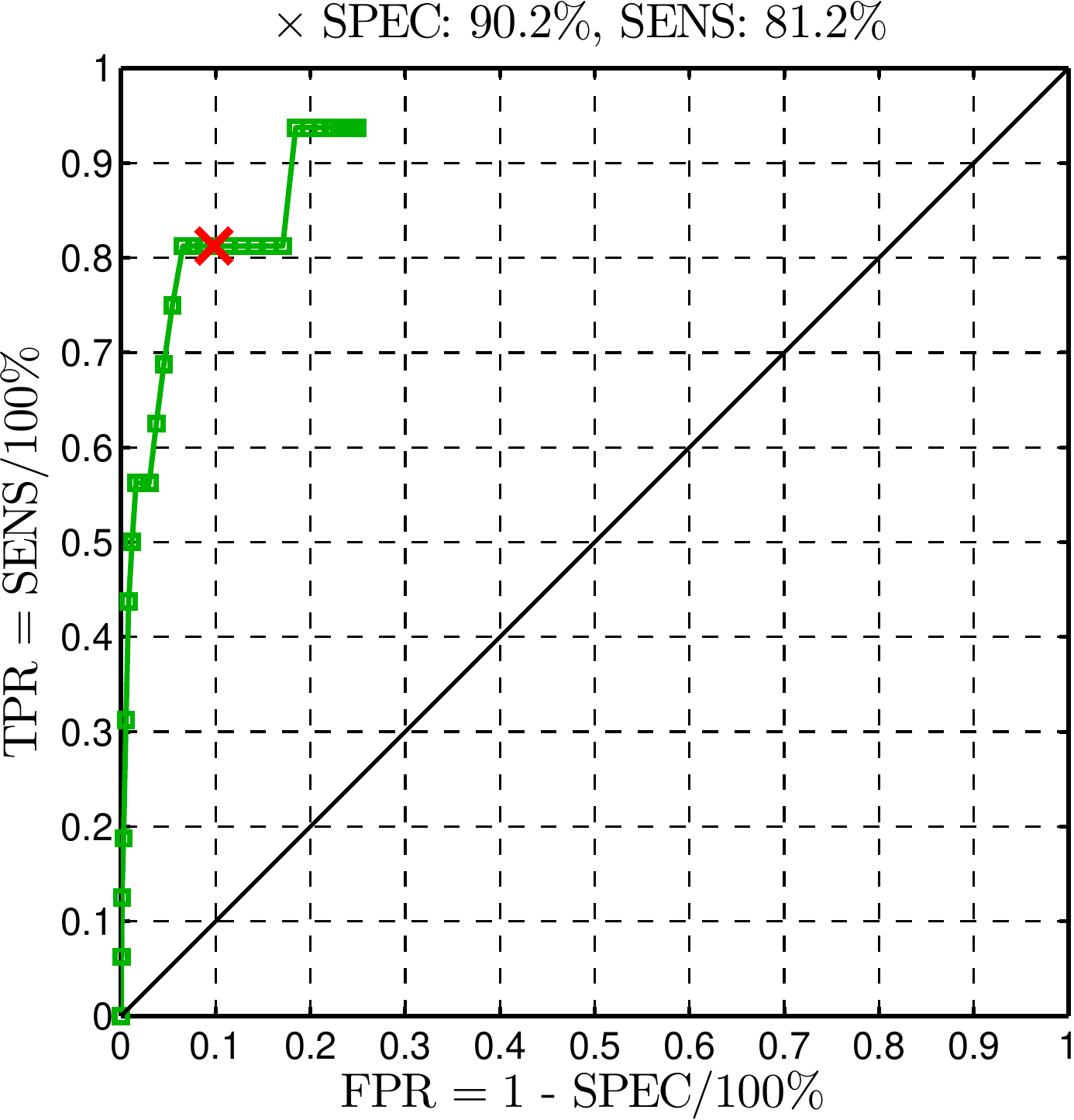} & \\
\centering (b) &
\end{tabular}
\caption{ROC curves for the binary classifier (\ref{eqn:binclass}):
(a) per frame basis, (b) per polyp basis.
Red $\times$ mark the points corresponding to $R_P = 37$.
No-discrimination lines are solid black.}
\label{fig:roc}
\end{figure}

\begin{table*}
\caption{Numbers $N_F^{(p)}$, $p=1,\ldots,N_S$ of frames in each of $N_S=16$ 
sequences of ``polyp'' frames. Flags $D^{(p)}$, $p=1,\ldots,N_S$ indicating 
if at least one frame in the corresponding sequence was classified as 
``polyp''. Per polyp sensitivity for $R_P = 37$ is $(13/16) \cdot 100\% = 81.25\%$.}
\centering
\begin{tabular}{l|c|c|c|c|c|c|c|c|c|c|c|c|c|c|c|c}
$N_F^{(p)}$ & 18 & 57 & 10 & ~7 & 11 & 11 & ~2 & ~3 & 17 & 32 & 12 & 15 & ~4 & ~7 & ~9 & 15 \\
\hline
$D^{(p)}$ & $\checkmark$ & $\checkmark$ & $\checkmark$ & $\checkmark$ & $\times$ & 
$\checkmark$ & $\checkmark$ & $\checkmark$ & $\checkmark$ & $\checkmark$ & $\times$ & $\times$ & 
$\checkmark$ & $\checkmark$ & $\checkmark$ & $\checkmark$
\end{tabular}
\label{tab:seq}
\end{table*}

We begin with a per frame ROC curve study, as described in section 
\ref{subsec:roc}. The ROC curve for the binary classifier (\ref{eqn:binclass}) 
is shown in Figure \ref{fig:roc} (a). Note that it does not go all the way 
to $(1,1)$. The reason for that is the use of pre-selection and combined
geometric criteria in steps \ref{step:text} and \ref{step:geom} of Algorithm 
\ref{alg:class} respectively. For the frames that do not satisfy those 
criteria we set $R_{max}=0$, but for plotting the ROC curve we only use 
positive $R_P$. However, this limitation is not important, since the portion 
of the ROC curve that is of most interest to us is the one corresponding to 
the small values of $FPR$. This is due to  the fact that an overwhelming 
majority of the frames in the endoscopy video  sequences are non-polyp 
frames (all of them for healthy patients). The frames that the algorithm 
labels as ``polyp'' have to be inspected manually by a doctor. Thus, to 
minimize the work that the doctor has to do, the specificity has to be high. 

According to our testing methodology, we set a target specificity at $90\%$.
This specificity is achieved on a training subset with a decision threshold
value of $R_P = 37$ resulting in the specificity of $91\%$. When the binary
classifier is applied to the whole data set with this threshold value, we
obtain the specificity of $90.2\%$ and the sensitivity of $47.4\%$, which
in shown in Figure \ref{fig:roc} (a). This is a good performance, 
especially considering the fact that it is achieved on a relatively
diverse data set. Moreover, such level of sensitivity for single frames 
can actually imply even better performance for video sequences, which is 
a more relevant way of evaluating the usefulness of the algorithm in 
the real clinical setting, as explained in section \ref{subsec:roc}.

For a per polyp study we first show the values of the detection flags 
$D^{(p)}$ in Table \ref{tab:seq} for all $N_S = 16$ polyp frame sequences. 
These values are obtained for the value of the discrimination threshold 
$R_P = 37$ as computed above. 
However, the sensitivity per polyp in this case is $81.25\%$, i.e. the 
algorithm correctly detects $13$ out of $16$ polyps in at least one frame 
of each corresponding sequence. Thus, as expected, the algorithm has a 
much better performance when a single polyp is present in a number of 
consecutive frames. This is further confirmed by Figure \ref{fig:roc} (b), 
where we show a per polyp ROC curve. In fact, we observe that we can 
obtain a specificity of $93.47\%$, while sill maintaining the same per 
polyp sensitivity of $81.25\%$ if we take $R_P = 40$.

By fixing the specificity at a high enough level we maintain control on
how many frames are to be inspected manually. An important measure that 
allows us to assess the burden of such manual inspection in real clinical
practice is the number of false positives and the false positive rate 
\emph{per patient}. We display these values in Table \ref{tab:falsepos}
for each of the five patients that our data comes from. In all cases we 
observe a massive reduction in the amount of frames that need to be 
inspected manually. For example, the training subset of our data set 
corresponding to Patient $4$ ($8567$ non-polyp frames) is reduced to
only $767$ frames that need to be inspected by a doctor.

\begin{table}
\caption{False positives per patient. For each of five patients the
number of normal (non-polyp) frames in the data set ($N_{norm}$), the 
number of false positives ($FPN$) and the false positive rate ($FPR$) 
are presented.}
\centering
\begin{tabular}{c|c|c|c}
Patient & $N_{norm}$ & $FPN$ & $FPR$ \\
\hline
1 & 422  &  34 & 8.0\% \\
\hline
2 & 1008 &  55 & 5.4\% \\
\hline
3 &	4666 & 671 & 14.3\% \\
\hline
4 & 8567 & 767 & 8.9\% \\
\hline
5 &	4075 & 310 & 7.6\% \\
\hline
Total & 18738 & 1837 & 9.8\%
\end{tabular}
\label{tab:falsepos}
\end{table}

\subsection{Robustness}
\label{subsec:resrobust}

\begin{table*}
\caption{Robustness study for Algorithm \ref{alg:class} with respect to the changes 
in the parameters. All values are given in percent. The base values $SPEC_{base}$, $SENS_{base}$
are in the ``Base'' row, the perturbed values $SPEC_{pert}$, $SENS_{pert}$ are in the rows below
along with the sensitivities $\delta(SPEC)$, $\delta(SENS)$, with each row corresponding to a
perturbation of a parameter indicated in the leftmost column.}
\centering
\begin{tabular}{c|l|l|l|l|l|l}
           & $SPEC$ & $\delta (SPEC)$ & $SENS$    & $\delta (SENS)$ & $SENS$    & $\delta (SENS)$ \\
					 &        &                 & per frame & per frame       & per polyp & per polyp \\
\hline
Base       & 92.2 & --   & 47.4 & --    & 81.2 & -- \\ 
\hline
$\sigma_1$ & 92.7 & 0.54 & 44.8 & 5.49  & 81.2 & 0 \\ 
\hline
$\sigma_2$ & 90.5 & 1.84 & 53.5 & 12.87 & 93.8 & 15.5 \\ 
\hline
$M_L$      & 92.3 & 0.11 & 47.0 & 0.84  & 81.2 & 0 \\ 
\hline 
$M_U$ 		 & 91.8 & 0.43 & 49.6 & 4.64  & 81.2 & 0 \\ 
\hline
$S_L$      & 92.2 & 0    & 47.4 & 0     & 81.2 & 0 \\ 
\hline
$S_U$      & 92.0 & 0.22 & 48.3 & 1.90  & 81.2 & 0 \\ 
\hline
$E_{max}$  & 91.7 & 0.54 & 48.7 & 2.74  & 81.2 & 0
\end{tabular}
\label{tab:sens}
\end{table*}

To assess the robustness of Algorithm \ref{alg:class} with respect to the 
changes in the numerical parameters, we perform the sensitivity calculations
as described in section \ref{subsec:robust}. The results of the sensitivity 
calculations are given in Table \ref{tab:sens}.
We observe that the algorithm is very robust with respect to all parameters,
with a possible slight exception of $\sigma_2$. For a $10\%$ relative 
perturbation of the parameters the sensitivity of $SPEC$ is less than $1\%$ 
for all cases except $\sigma_2$, for which it is $1.8\%$. The sensitivity of 
per frame $SENS$ is larger, but still it is around $5\%$ or below for all 
parameters except for $\sigma_2$. The per polyp $SENS$ is even more robust. 
The $10\%$ parameter perturbation does not affect it at all with a same single 
exception. This is a remarkable display of robustness, considering that the 
algorithm is a highly non-linear procedure with lots of conditional branching 
based on thresholds.

The behavior in the single exceptional case (sensitivity with respect to 
$\sigma_2$) can be explained by the fact that $\sigma_2$ directly affects the
maximal size of features in the mid-pass filtered frame. Thus, the increase in 
$\sigma_2$ leads to the increase in $R_{max}$. Since Algorithm \ref{alg:class}
identifies large values of $R_{max}$ with polyps, this should lead to increased
$SENS$, which we observe in Table \ref{tab:sens} for both per frame ($53.5\%$ 
perturbed against $47.4\%$ base) and per polyp ($93.8\%$ perturbed against 
$81.2\%$ base) basis. Even then, the relative sensitivity remains comparable
with the size of the relative perturbation in $\sigma_2$ ($12.8\%$ per frame,
$15.5\%$ per polyp against $10\%$).

\section{Discussion}
\label{sec:discuss}

In this paper we developed an algorithm for automated detection of polyps in the 
images captured by a capsule endoscope. The problem of polyp detection is quite
challenging due to a multitude of factors. These include the presence of trash 
liquids and bubbles, vignetting due to the use of a non-uniform light source,
high variability of possible polyp shapes and the lack of a clear cut between
the geometry of the polyps and the folds of a healthy mucosal tissue. We attempt  
to overcome these issues by utilizing both the texture information and the 
geometrical information present in the frame to obtain a binary classification
algorithm with pre-selection. 

We perform a thorough statistical testing of the algorithm on a rich data set 
to ensure its good performance in realistic conditions. The algorithm demonstrates
high per polyp sensitivity and, equally importantly, displays a high per
patient specificity, i.e. a consistently low false positive rate per individual
patient. Such behavior is desirable as one of the main goals of automated polyp 
detection is to drastically decrease the amount of video frames that require 
manual inspection. While our approach is by no means an ultimate solution of the 
automated polyp detection problem, the achieved performance makes this work an
important step towards a fully automated polyp detection procedure.

Throughout the paper we identified some directions of future research and the
possible areas of improvement of the algorithm. We summarize them in the list
below. 

\begin{itemize}
\item Currently, the frame is converted to grayscale before processing, so the
algorithm only utilizes the information about the texture and the geometry. 
It would be beneficial to also utilize the color information present in the
frame. For example, the amount of red color can point to polyps that are highly
vascularized.
\item The effectiveness of our algorithm lies partially in the use of a 
pre-selection criterion. It is based on an idea of filtering out the 
\emph{non-informative} frames without any further consideration. While using
the texture content is a simple and robust procedure, more complicated 
pre-selection approaches can be used, e.g. from \cite{liu2013robust, oh2007informative}.
In particular, in \cite{oh2007informative} the use of a Discrete Fourier 
Transform is proposed for frame classification as informative/non-informative.
Another procedure from \cite{oh2007informative} that can be used to improve 
our pre-selection criterion is the specular reflection detection. It should
be particularly effective in filtering out the frames with bubbles, since those
typically produce strong specular reflections.
\item The algorithm relies on a number of parameters. For the purpose of this
study the values of most parameters were chosen manually. While the sensitivity 
computations in section \ref{subsec:robust} demonstrate the robustness of the
algorithm with respect to the changes in the parameter values, one may use an 
automated calibration procedure to choose these parameters. Developing
such a procedure remains a topic of future research.
\item A binary classifier for polyp detection is straightforward to implement 
and its performance can be easily assessed with the help of ROC curves. 
However, using more advanced machine learning and classification techniques 
may improve the detection of polyps. For example, support vector machines
\cite{cortes1995support} were used in a polyp detection method based on a 
segmentation approach \cite{condessa2012segmentation}. Alternatively, one
can use a random forest \cite{breiman2001random, ho1998random} to construct 
a classifier. 
\item In order to properly assess the size and shape of the protrusions, the 
algorithm should be able to correctly infer the actual height map of the object 
from the intensity information in the image. This presents a particular challenge
when an object of interest is located in the dark section of the image. Using
the mid-pass filtering provides an adequate solution to inferring the height
map from the image. Nevertheless, we would like to investigate possible alternatives
to the approach used here.
\item Even if an algorithm perfectly detects polyp frames in a video sequence, 
it does not detect the actual location of a polyp in a colon. This problem is
particularly exacerbated in capsule colonoscopy due to the highly irregular 
motion of the capsule. One way to overcome this issue is to try to reconstruct 
the capsule's motion from the changes in the subsequent video frames. Alternatively,
one may employ the approaches used in conventional colonoscopy. For example, 
in \cite{liu2013optical} a procedure is proposed for co-alignment of the optical 
colonoscopy video with a virtual colonoscopy from an X-ray CT exam. Tracking the 
movement of a capsule remains a topic of our research, whether by combining 
with other imaging modalities or by extracting the motion information 
directly from the video sequence.
\end{itemize}

\section*{Acknowledgments}

This work was partially supported by CoLab, the UT Austin $|$ Portugal 
International Collaboratory for Emerging Technologies 
(\url{http://utaustinportugal.org}), project UTAustin/MAT/0009/2008 
and also by project PTDC/MATNAN/0593/2012, 
by CMUC and FCT (Portugal, through European program COMPETE/FEDER and project 
PEst-C/MAT/UI0324/2011). The work of Y.-H. R. Tsai was partially supported
by Moncrief Grand Challenge Award and by the National Science Foundation grant 
DMS-1318975.

The authors thank the anonymous referees for valuable comments and suggestions
that helped to improve the manuscript.
 
\bibliography{biblio-rev1} 
\bibliographystyle{siam}

\end{document}